\DeclareUnicodeCharacter{202F}{\,}
\documentclass[11pt,oneside,a4paper,onecolumn]{elsarticle}

\usepackage{supertabular}
\usepackage{longtable,tabularx}
\usepackage{multirow}
\usepackage{times}
\usepackage{fullpage}

\usepackage{gensymb}
\usepackage{tikz}
\usepackage{tikzscale}
\usepackage{alltt}
\usepackage{upquote}
\usepackage{float}
\usepackage{makecell}
\usepackage{array}
\usetikzlibrary{positioning}
\usepackage{graphicx}
\usepackage[toc,page]{appendix}
\usepackage[utf8]{inputenc}
\usepackage{amsmath,mathtools}
\usetikzlibrary{calc,decorations.markings,fit,shapes,chains,arrows}
\usepackage{amsthm}
\usepackage{amsfonts}
\usepackage{amssymb}
\usepackage{enumitem}
\usepackage{caption}
\usepackage{subcaption}
\usepackage{scalerel}
\usepackage{algorithm}
\usepackage{algorithmic}

\usepackage{multicol}
\usepackage{tikz}
\usetikzlibrary{arrows.meta,positioning,fit}

\usepackage{multirow}
\usepackage{url}
\usepackage{booktabs}
\usepackage{nicefrac}

\usepackage{soul}
\usepackage{slashed}
\usepackage{lineno}

\usepackage{xcolor}
\modulolinenumbers[5]

\usepackage[final]{hyperref} 
\hypersetup{
	colorlinks=true,       
	linkcolor=blue,        
	citecolor=blue,        
	filecolor=magenta,     
	urlcolor=blue         
}

\newcolumntype{M}[1]{>{\centering\arraybackslash}m{#1}}
\newcolumntype{N}{@{}m{0pt}@{}}

\tikzset{>=latex}
\def\@author#1{\g@addto@macro\elsauthors{\normalsize%
    \def\baselinestretch{1}%
    \upshape\authorsep#1\unskip\textsuperscript{%
      \ifx\@fnmark\@empty\else\unskip\sep\@fnmark\let\sep=,\fi
      \ifx\@corref\@empty\else\unskip\sep\@corref\let\sep=,\fi
      }%
    \def\authorsep{\unskip,\space}%
    \global\let\@fnmark\@empty
    \global\let\@corref\@empty
    \global\let\sep\@empty}%
    \@eadauthor={#1}
}

\graphicspath{{Images/}}

\begin{document}
\begin{frontmatter}
\title{A Physics-Guided Probabilistic Surrogate Modeling Framework for Digital Twins of Underwater Radiated Noise}

\author[ubc]{Indu Kant Deo\corref{cor1}}
\ead{indukant@mail.ubc.ca}

\author[ubc]{Akash Venkateshwaran}
\ead{akashv22@student.ubc.ca}

\author[ubc]{Rajeev K. Jaiman}
\ead{rjaiman@mail.ubc.ca}
\cortext[cor1]{Corresponding author}
\address[ubc]{Department of Mechanical Engineering, The University of British Columbia, Vancouver, BC V6T 1Z4}

\begin{abstract}
Ship traffic is an increasing source of underwater radiated noise in coastal waters, motivating real-time digital twins of ocean acoustics for operational noise mitigation. We present a physics-guided probabilistic framework to predict three-dimensional transmission loss in realistic ocean environments. As a case study, we consider the Salish Sea along shipping routes from the Pacific Ocean to the Port of Vancouver. A dataset of over 30 million source–receiver pairs was generated with a Gaussian beam solver across seasonal sound speed profiles and one-third-octave frequency bands spanning 12.5\,Hz to 8\,kHz. 
We first assess sparse variational Gaussian processes (SVGP) and then incorporate physics-based mean functions combining spherical spreading with frequency-dependent absorption. To capture nonlinear effects, we examine deep sigma-point processes and stochastic variational deep kernel learning. The final framework integrates four components: (i) a learnable physics-informed mean that represents dominant propagation trends, (ii) a convolutional encoder for bathymetry along the source–receiver track, (iii) a neural encoder for source, receiver, and frequency coordinates, and (iv) a residual SVGP layer that provides calibrated predictive uncertainty. This probabilistic digital twin facilitates the construction of sound-exposure bounds and worst-case scenarios for received levels.
We further demonstrate the application of the framework to ship speed optimization, where predicted transmission loss combined with near-field source models provides sound exposure level estimates for minimizing acoustic impacts on marine mammals. The proposed framework advances uncertainty-aware digital twins for ocean acoustics and illustrates how physics-guided machine learning can support sustainable maritime operations.

\smallskip
\smallskip

  \end{abstract}
\end{frontmatter}


\section{Introduction}
The marine industry represents nearly 75\% of global freight transport and has expanded substantially since the early 2000s. The increase in maritime traffic has been directly linked to elevated ambient low-frequency noise levels (10 to 100 Hz) in the world's oceans, studies estimating increases of up to 3 dB per decade \cite{mcdonald2006increases,andrew2011long}. Because sound propagates more efficiently in water than in air, it is the primary medium for communication, navigation, and prey detection among marine species such as whales, dolphins, and fish \cite{richardson2013marine}. Anthropogenic noise has measurable effects on marine life, ranging from behavioral changes to mortality, depending on intensity, frequency, and proximity to the source. Addressing ship noise has therefore become an important objective for regulators and the scientific community in the conservation of marine ecosystems.


Real-time prediction of underwater acoustic fields is essential for mitigating underwater radiated noise (URN). Traditional approaches to modeling sound propagation typically employ reduced representations of the wave equation to achieve computational efficiency. Common formulations include ray-tracing techniques \cite{porter1987gaussian,porter2019beam,jensen2011computational}, parabolic equation methods \cite{collins1989applications,tappert2005parabolic,collins1993split}, and normal mode theory \cite{williams1970normal,chapman1990normal}. These models explicitly incorporate environmental parameters such as spatially varying sound-speed profiles, bathymetry, and seabed properties to resolve the acoustic field. While they can yield accurate predictions in controlled settings, their use in operational contexts is limited by the difficulty of assimilating heterogeneous data sources (e.g., sensor measurements and simulations) and by the computational cost, which restricts their applicability in time-critical tasks such as ship routing and noise mitigation.

To address the limitations of physics-based models, data-driven approaches have been investigated for predicting underwater acoustic fields across diverse environmental conditions. These methods are designed to generalize over varying oceanographic inputs and provide computationally efficient alternatives to traditional solvers. With the availability of large-scale acoustic datasets, machine learning techniques have been used to learn mappings from input conditions to acoustic fields in real time, thereby reducing the reliance on explicit environmental modeling \cite{irfan2021deepship,wang2016machine}. Approaches such as Gaussian process regression \cite{bernardo1998regression,caviedes2021gaussian,deo2025data} and deep neural networks \cite{liu2017survey,deo2022predicting,mallik2022predicting,deo2024continual,deo2024predicting,mccarthy2023reduced,deo2025predicting} can approximate complex acoustic fields when sufficient training data and model capacity are available. Physics-guided deep learning architectures that embed physical constraints within the network structure have also been applied to underwater acoustic propagation \cite{Deo_2025,du2023research,huang2024broadband,xi2024fem}. However, the temporal variability of ocean conditions necessitates frequent model updates to maintain predictive accuracy, motivating their integration within digital twin frameworks for adaptive and real-time acoustic field estimation.

\begin{figure}[htbp]
    \centering
    \includegraphics[width=1.0\linewidth]{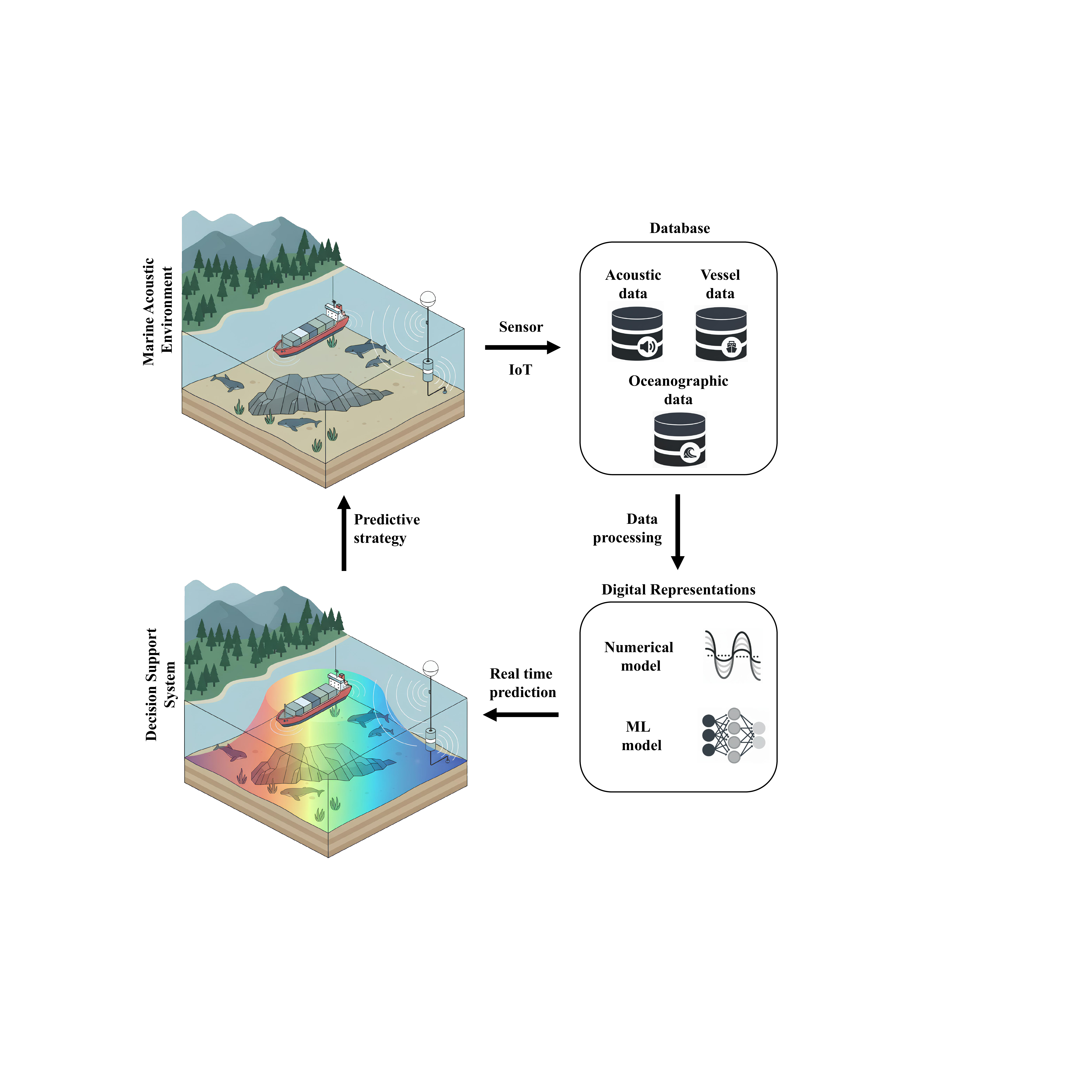}
    \caption{Illustration of the digital twin framework for ocean soundscape modeling, integrating ship tracks and hydrophone data with physics-based and machine learning solvers for real time acoustic prediction.}
    \label{fig:digital_twin_tc}
\end{figure}

A digital twin is a virtual representation of a physical system that is continuously updated with real-world data \cite{xiu2025computational}. It combines physics-based simulations, sensor observations, and machine learning models to characterize the current state of the system and forecast its evolution \cite{henneking2025goal}. In underwater acoustics, digital twin frameworks have been developed as practical tools for real-time prediction of the ocean soundscape \cite{fuller2021real,abbott2022digital,egu2020digital}. These frameworks integrate data streams such as hydrophone recordings, environmental measurements, and vessel locations to generate acoustic field predictions in real time. As new observations become available, the models are updated to reflect changes in oceanographic and anthropogenic conditions. The general structure of such a framework is shown in Fig.~\ref{fig:digital_twin_tc}. In this work, we build on this structure to develop a physics-guided probabilistic digital twin for underwater acoustic propagation, enabling calibrated uncertainty quantification and real-time decision support.

\begin{figure}[h]
    \centering
    \includegraphics[width=0.99\linewidth,keepaspectratio]{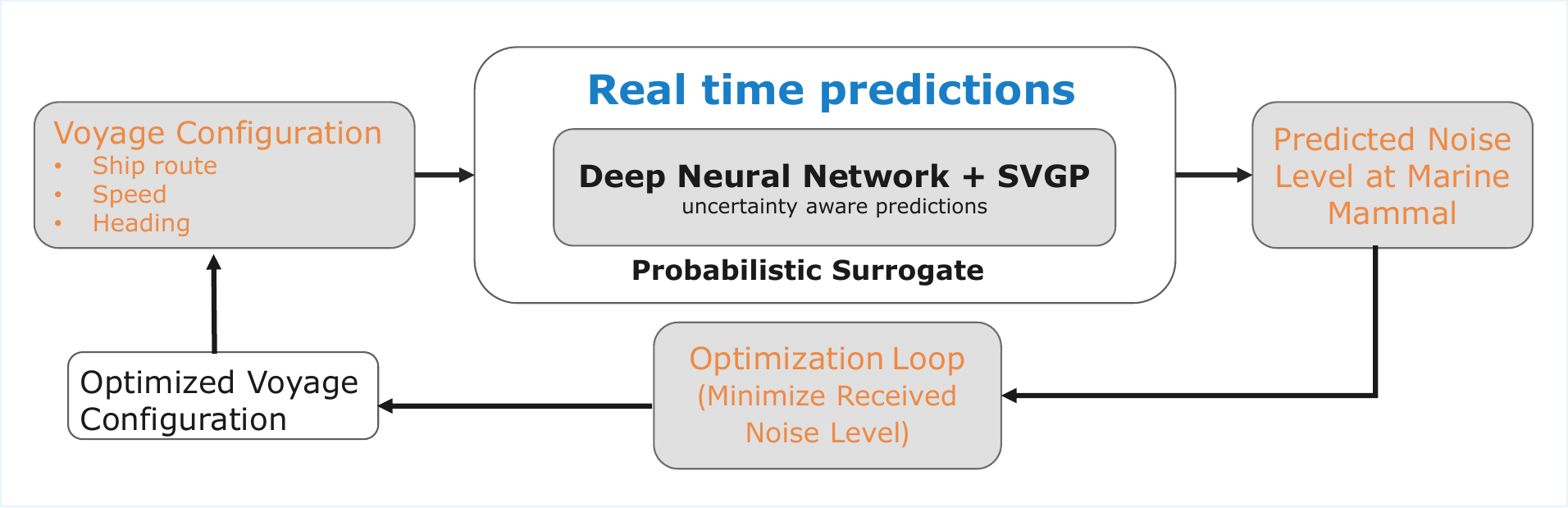}
    \caption{Digital twin framework for real-time voyage optimization, integrating uncertainty-aware acoustic field prediction with vessel routing and speed control.}
    \label{fig:digital_twin_optimization}
\end{figure}

In the context of marine conservation, digital twins enable real-time prediction of underwater noise in ecologically sensitive areas, supporting mitigation strategies such as vessel speed reduction and routing adjustments. Unlike static models, digital twins incorporate feedback mechanisms that improve prediction quality over time, making them suitable for decision-making in dynamic environments. The prediction of acoustic fields in coastal waters, however, often relies on numerical solutions of partial differential equations or on data-driven surrogates such as convolutional neural networks, which have shown success under constrained conditions \cite{venkateshwaran2024multi,venkateshwaran2024multiOMAE,venkateshwaran2025mute}. A key limitation of purely data-driven approaches is the lack of quantified uncertainty, which reduces the reliability of predictions in extreme or previously unobserved environments. Figure~\ref{fig:digital_twin_optimization} shows a digital twin framework that integrates real-time acoustic prediction with voyage optimization to minimize received noise levels. This motivates the development of physics-guided probabilistic surrogates with calibrated uncertainty for robust digital twin applications.

Gaussian processes (GP) provide a principled Bayesian framework for function approximation and uncertainty quantification, offering closed-form posterior distributions for predictions conditioned on training data \cite{mackay1998introduction}. Their flexibility arises from the specification of a kernel function, which encodes prior assumptions about smoothness, periodicity, or other structural properties of the target function. However, the cubic computational complexity of the number of training points renders standard GP impractical for large-scale datasets common in ocean acoustics.
To address this issue, sparse approximations were introduced, wherein a reduced set of inducing points is used to approximate the full covariance structure. Variational formulations, such as the stochastic variational Gaussian process (SVGP), extend this approach by optimizing a tractable evidence lower bound, allowing scalability to millions of training points through minibatch stochastic gradient descent \cite{hensman2015scalable,blei2017variational}. These methods preserve calibrated uncertainty estimates while achieving substantial computational gains, making them attractive for real-time or near-real-time applications.

Beyond scalability, there is interest in enhancing the expressiveness of GP models. One approach incorporates physics-informed mean functions or kernels that embed domain knowledge, thereby improving prediction and reducing bias in data-sparse regimes. Another direction is hierarchical or deep GP formulations, where the outputs of one GP layer serve as inputs to another. Among these, deep sigma-point processes (DSPPs) have been proposed to capture multi-scale nonlinearities by propagating sigma points through compositional GP layers \cite{jankowiak2020deep,su2023deep}. Although promising for representing complex dependencies, DSPPs often exhibit training instability and remain computationally demanding for high-dimensional input. In parallel, hybrid models such as stochastic variational deep kernel learning (SVDKL) integrate deep neural networks with variationally trained GPs. In this formulation, the neural network acts as a feature extractor that maps high-dimensional inputs into a latent representation, on which a GP with inducing points is placed. This allows SVDKL to take advantage of the representation power of neural networks while retaining the calibrated uncertainty estimates of GPs \cite{wilson2016stochastic,wilson2016deep}. Applications have shown that SVDKL can model complex, non-stationary data distributions that are intractable for conventional kernels. However, despite these advances, both DSPPs and SVDKL remain limited in coastal acoustics, where spatially varying bathymetry, frequency dependence, and seasonal sound-speed profiles introduce multiscale heterogeneity that is not easily captured by standard kernels or learned feature mappings. This motivates the development of a physics-guided probabilistic GP framework tailored for coastal underwater acoustic propagation.

GP-based frameworks, including sparse variational, hierarchical, and hybrid deep kernel models, provide building blocks for uncertainty-aware learning. However, they remain limited in scalability, robustness, and interpretability when applied to 3D range-dependent acoustic propagation. These limitations motivate a probabilistic digital twin that integrates physics-informed structure, bathymetry-aware encoding, and scalable variational inference for real-time prediction in complex marine environments. The model is trained on more than 30 million Bellhop3D-simulated transmission loss data points across frequency bands from 12.5 Hz to 8 kHz, sampled over source–receiver locations and seasonal sound-speed profiles. The surrogate architecture includes four components: a physics-informed mean function for spherical spreading with Thorp absorption, a one-dimensional convolutional bathymetry encoder, a neural encoder for source–receiver coordinates and frequency, and a residual SVGP layer for calibrated uncertainty estimates.

This formulation preserves physical interpretability, achieves computational efficiency, and provides robust uncertainty quantification. We further demonstrate the application of the probabilistic digital twin to operational decision-making, with a focus on ship speed optimization to minimize sound exposure levels (SEL) to marine mammals, a critical requirement for sustainable marine operations.
In summary, the main contributions of this work are as follows.
\begin{enumerate}
\item Integration of physics-based mean functions and learned encoders with a variational GP residual, providing robustness and calibrated uncertainty across seasonal sound-speed profiles and range-dependent bathymetry.
\item Development of a scalable surrogate trained on more than 30 million Bellhop3D simulations, achieving sub-second inference and errors below 1 dB relative to high-fidelity models.
\item An end-to-end digital twin framework that couples real-time acoustic field prediction with voyage optimization for operational noise mitigation in complex coastal environments.
\end{enumerate}

The remainder of this paper is organized as follows. Section~\ref{sec:problem_formulation} introduces the mathematical preliminaries and the probabilistic digital twin framework. Section~\ref{sec:method} describes the neural encoders and the SVGP residual component that form the core of the architecture. Section~\ref{sec:training} outlines the simulation setup, the generation of ground-truth transmission loss data, and the training procedure. Section~\ref{sec:results} presents numerical results demonstrating the predictive performance of the framework for three-dimensional transmission loss in a coastal shipping corridor near the Port of Vancouver. Section~\ref{sec:conclusion} concludes with a summary of the main findings and future research directions.

\section{Mathematical Background}
\label{sec:problem_formulation}
Underwater noise transmission is governed by the propagation of acoustic pressure waves generated by sound sources. By linearizing the hydrodynamic equations and applying the equation of state, one obtains a wave equation that describes small pressure perturbations in the ocean. Because oceanographic processes evolve on much longer timescales than acoustic propagation, the medium properties, including the squared sound speed $c^2$, are often treated as temporally invariant \cite{jensen2011computational}. Under this assumption, the homogeneous acoustic wave equation for pressure perturbation is
\begin{equation}
\label{eq:pressure_waveEQN}
\rho_{0} \nabla \cdot \left(\frac{1}{\rho_{0}} \nabla p^{'} \right) - \frac{1}{c^2} \frac{\partial^2 p^{'}}{\partial t^2} = 0,
\end{equation}
where $p^{'}$ denotes the pressure perturbation and $\rho_{0}$ is the local time-averaged density. 

For spatially uniform density, Eq.~\ref{eq:pressure_waveEQN} reduces to the standard homogeneous wave equation
\begin{equation}
\label{eq:std_pressure_waveEQN}
\nabla^2 p - \frac{1}{c^2} \frac{\partial^2 p}{\partial t^2} = 0 ,
\end{equation}
where the prime notation for pressure perturbations has been omitted. Acoustic sources are incorporated by introducing a forcing term on the right-hand side of Eq.~\ref{eq:std_pressure_waveEQN}, typically modeled as a space–time distribution function.
Since the coefficients of the operators in Eq.~\ref{eq:std_pressure_waveEQN} are independent of time, the equation can be transformed into the frequency domain using a Fourier transform in time. This reduction leads to the inhomogeneous Helmholtz equation in Cartesian coordinates $\mathbf{x}=(x,y,z)$,
\begin{equation}
\label{eq:helmohotz_eqn}
\nabla^2 p + \frac{\omega^2}{c^2(\mathbf{x})} p = -\delta(\mathbf{x} - \mathbf{x}_0),
\end{equation}
where $\omega$ is the angular frequency of the point source located at $\mathbf{x}_0$, and $c(\mathbf{x})$ represents the spatially varying sound speed field.
To complete the formulation, appropriate boundary conditions must be specified on the computational domain. The general boundary condition can be expressed as
\begin{equation}
\label{eq:helmohotz_eqn_BC}
\alpha , p(\partial\mathbf{x}; \omega) + \beta , \frac{\partial p(\partial\mathbf{x}; \omega)}{\partial \mathbf{n}} = g(\partial\mathbf{x}; \omega),
\end{equation}
where $\partial\mathbf{x}$ denotes the boundary of the domain, $\mathbf{n}$ is the outward normal vector pointing into the ocean domain, and $\alpha$ and $\beta$ are scalar coefficients. By choosing appropriate values of $\alpha$ and $\beta$, Eq.~\ref{eq:helmohotz_eqn_BC} can be specialized to represent different physical scenarios, such as pressure-release boundaries at the sea surface, rigid seabed boundaries, or absorbing boundary conditions that approximate outgoing waves at open ocean boundaries.

The acoustic energy or intensity $\left(I(z,R), ; I \propto p^2\right)$ decreases with increasing distance from the source, a phenomenon referred to as transmission loss (TL). This quantity is typically expressed in decibels by comparing the acoustic pressure $p(\mathbf{X}_{rcv},\omega_{src})$ at a receiver location $(\mathbf{X}_{rcv})$ with the reference signal $p_0$, which corresponds to the pressure emitted by the source scaled to one meter:
\begin{equation}
\label{eq:TL_comp}
\mathrm{TL} (R,z) = -10 \log \frac{I(\mathbf{X}_{rcv},\omega_{src})}{I_0} = - 20 \log \frac{|p(\mathbf{X}_{rcv},\omega_{src})|}{|p_0|} \quad \text{(dB re 1 m)}.
\end{equation}
Transmission loss quantifies the reduction in far-field sound pressure level from an underwater radiated noise (URN) source. In practice, TL arises from multiple processes, including geometric spreading, absorption, scattering, reflection, refraction, and other environmental losses.

The dominant contributions to transmission loss are geometric spreading and absorption. Geometric spreading represents the spatial divergence of the acoustic wavefront and is modeled as a logarithmic decay with range:
\begin{equation}
\label{eq:geo_spread}
\mathrm{TL}_{\text{spreading}} = A \log_{10}(R),  \text{where}   \begin{cases}
    A = 10, & \text{Cylindrical spreading} \\
    A = 20, & \text{Spherical spreading}
    \end{cases}
\end{equation}
where $R$ is the source–receiver distance. Absorption accounts for the frequency-dependent attenuation of acoustic energy as it propagates through water and is typically modeled as a linear function of distance:
\begin{equation}
\label{eq:absorp}
\mathrm{TL}_{\text{absorption}} = \alpha(\omega) , R,
\end{equation}
where $\alpha(\omega)$ is the volume absorption coefficient. This coefficient is commonly estimated using Thorp’s empirical formula, which accounts for the molecular relaxation effects of boric acid and magnesium sulfate, as well as the viscosity of pure water. Additional processes such as seabed interaction, surface reflection, scattering, and refraction due to sound-speed gradients also contribute to transmission loss. These effects are strongly environment dependent and are typically resolved only by high-fidelity solvers such as parabolic equation models, Gaussian beam tracing, or normal mode theory. In the present framework, spreading and absorption are included analytically, while the remaining processes are represented by the GP residual.

In the present work, we model the transmission loss as a sum of two components: a physics-informed mean function that explicitly accounts for geometric spreading and absorption (Eqs.~\ref{eq:geo_spread}–\ref{eq:absorp}), and a data-driven residual term that represents unresolved processes such as scattering, reflection, and refraction. The residual is modeled using a stochastic variational Gaussian process framework, which scales the representation power of Gaussian processes to a large dataset. This hybrid formulation allows us to retain physical interpretability for the dominant mechanisms while leveraging data-driven learning to capture complex environmental interactions beyond the reach of simplified analytical models.


\section{Proposed Surrogate Modeling Framework}\label{sec:method}
\begin{figure}[htbp]
    \centering
    \makebox[\textwidth][c]{%
    \includegraphics[width=1.3\linewidth]{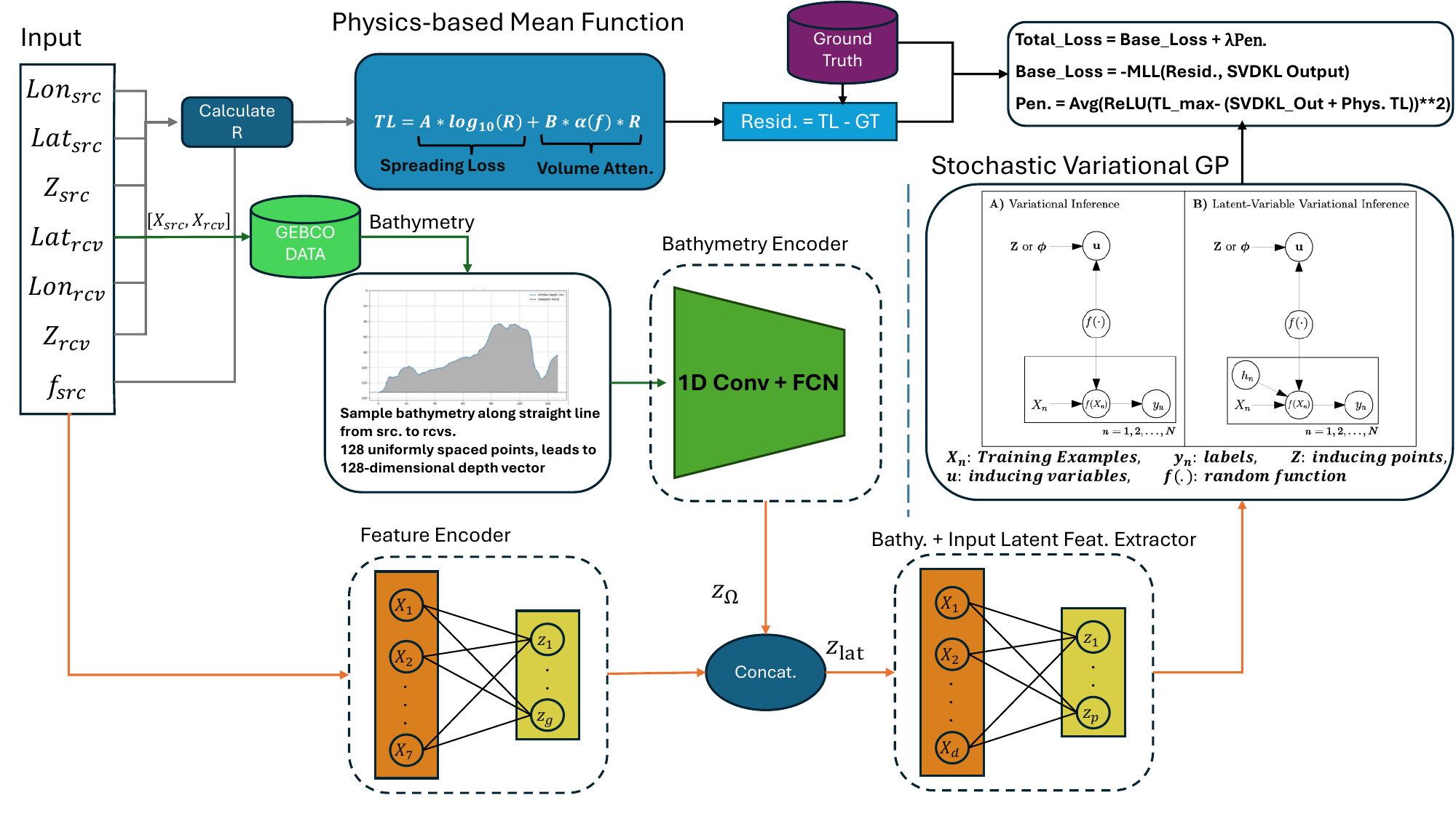}
    }
    \caption{Architecture of the probabilistic digital twin. The model integrates a physics-informed mean with neural encoders for bathymetry and geometry, followed by an SVGP residual head for uncertainty-aware prediction.}
    \label{fig:gp_architecture}
\end{figure}

The proposed modeling framework combines physics-based formulations with scalable probabilistic machine learning to predict transmission loss. The architecture, illustrated in Fig.~\ref{fig:gp_architecture}, consists of three main components: (i) a physics-informed mean function that encodes geometric spreading and absorption, (ii) neural encoders for bathymetry and geometric features, and (iii) a stochastic variational Gaussian process residual component that captures unresolved variability and provides calibrated uncertainty estimates. In the following subsections, we describe each component in detail.

\subsection{Physics-Informed Mean Function}
To retain physical interpretability, the surrogate is constructed around a physics-informed mean function that captures the dominant mechanisms of transmission loss. Geometric spreading and frequency-dependent absorption are modeled analytically, while remaining processes such as scattering, refraction, and seabed interaction are represented by a data-driven residual. This separation embeds well-established physical laws while allocating complexity to the probabilistic component.

Transmission loss is decomposed into dominant analytic contributions and a residual component:  
\begin{equation}
    \mathrm{TL}(R,f) = \underbrace{A \log_{10}(R) + B \, \alpha(f) R}_{\text{Physics-informed mean}} + r_{\theta}(R,f,\mathbf{X_s}, \mathbf{X_r}, \Omega),
\end{equation}
where $R$ is the source--receiver distance, $f$ is frequency, and $\alpha(f)$ is the frequency-dependent absorption coefficient. Parameters $A$ and $B$ are learnable scale factors. The residual $r_{\theta}$ captures processes not explained by simple spreading and absorption, such as scattering, refraction, and seabed interaction. The absorption coefficient \(\alpha(f)\) follows Thorp’s empirical formula \cite{fisher1977sound} given as:
\begin{equation}
\alpha(f) = \frac{0.11\,f^{2}}{1+f^{2}} + \frac{44\,f^{2}}{4100 + f^{2}} + 2.75\times 10^{-4}\,f^{2} + 0.003,
\end{equation}
where \(f\) is in kHz and \(\alpha(f)\) is in dB/km.
\subsection{Neural Encoders \label{sec:neural_enc}}

\begin{figure}[htbp]
    \centering
    \includegraphics[width=0.99\linewidth]{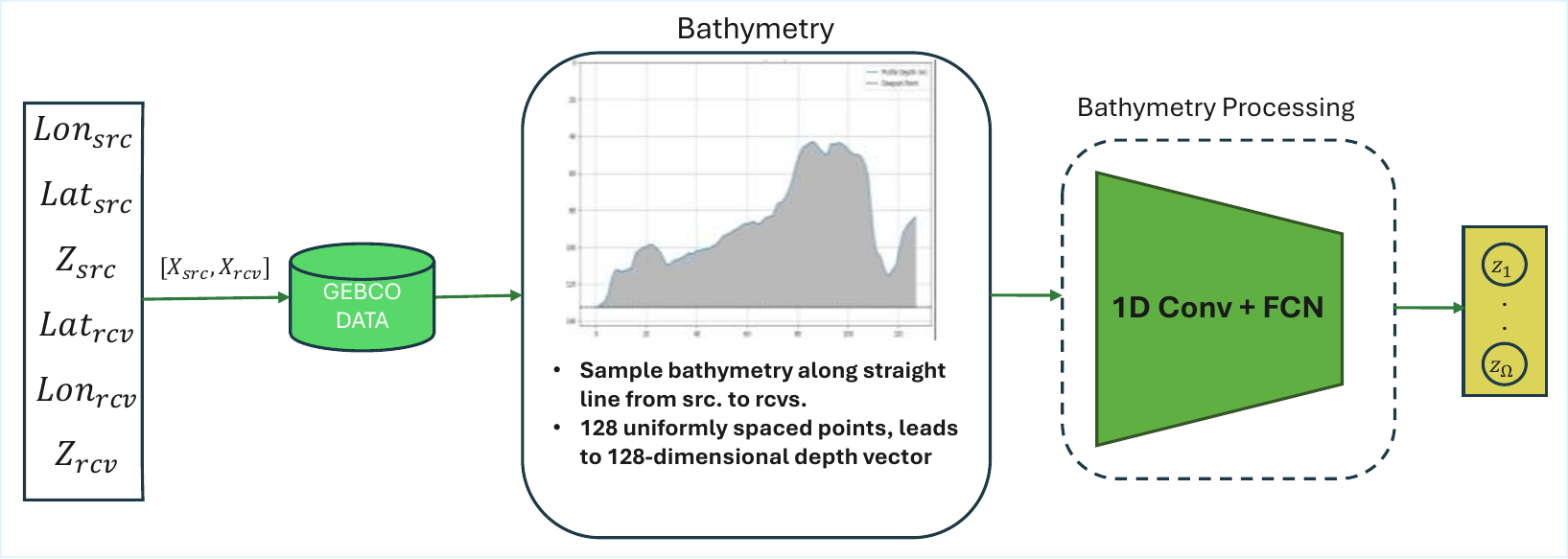}
    \caption{Pipeline of the bathymetry encoder. The 128-point bathymetric profile $\Omega$ is processed through convolutional layers, pooling, flattening, and fully connected layers to produce the embedding vector $\mathbf{z}_\Omega$.}
    \label{fig:bathy_encoder_arch}
\end{figure}

\paragraph*{Bathymetry Encoder}  
Bathymetry strongly influences transmission loss through seabed interaction and refraction, motivating an explicit encoder for this feature. The bathymetry profile along the source--receiver path is sampled at $128$ uniformly spaced points from the GEBCO dataset \cite{GEBCO_2024}, yielding a discrete vector
\[
\Omega = \{d_1, d_2, \dots, d_{128}\}, \quad d_i \in \mathbb{R},
\]
where each $d_i$ corresponds to the seabed depth at one sampling location. The bathymetry encoder maps $\Omega \in \mathbb{R}^{B \times K}$, with $K=128$ and batch size $B$, to a latent representation $\mathbf{z}_\Omega \in \mathbb{R}^{B \times d_\Omega}$. The encoding is implemented using a one-dimensional convolutional neural network followed by pooling and fully connected layers. At an abstract level, this mapping can be written as
\begin{equation}
    \mathbf{z}_\Omega = f_{\text{MLP}}\!\left(\text{Flatten}\!\left(\text{Pool}\!\left(f_{\text{CNN}}(\Omega)\right)\right)\right),
\end{equation}
where $f_{\text{CNN}}$ is a stack of Conv1D layers with SiLU activations and batch normalization, $\text{Pool}(\cdot)$ is adaptive average pooling, and $f_{\text{MLP}}$ is a multi-layer perceptron producing the embedding. The detailed specification of the encoder is summarized in Table~\ref{tab:bathy_encoder}, and the complete pipeline is illustrated in Fig.~\ref{fig:bathy_encoder_arch}.

\begin{table}[htbp]
\centering
\caption{Architecture of the Bathymetry Encoder. Input profile has dimension $(B,K)$ with $K=128$, output embedding has dimension $(B,\text{emb\_dim})$.}
\label{tab:bathy_encoder}
\begin{tabular}{llll}
\hline
\textbf{Stage} & \textbf{Layer} & \textbf{Output Shape} & \textbf{Details} \\
\hline
Input  & Bathymetry profile & $(B, K)$ & Normalized to $[0,1]$ \\
       & Unsqueeze channel  & $(B,1,K)$ & Add channel dimension \\[3pt]
\hline
\multirow{4}{*}{Conv block} 
 & Conv1D + SiLU + BN & $(B,8,K)$   & 1$\to$8 channels, kernel=5, padding=2 \\
 & Conv1D + SiLU + BN & $(B,16,K)$  & 8$\to$16 channels, kernel=5, padding=2 \\
 & Conv1D + SiLU + BN & $(B,32,K)$  & 16$\to$32 channels, kernel=5, padding=2 \\
 & AdaptiveAvgPool1D  & $(B,32,16)$ & Temporal dimension reduced to 16 \\[3pt]
\hline
\multirow{4}{*}{Head (MLP)} 
 & Flatten             & $(B,512)$   & Flatten $(32 \times 16)$ \\
 & Linear + SiLU + BN  & $(B,256)$   & Fully connected, 512$\to$256 \\
 & Linear + SiLU + BN  & $(B,128)$   & 256$\to$128 \\
 & Linear + SiLU + BN  & $(B,64)$    & 128$\to$64 \\
 & Linear              & $(B,\text{emb\_dim})$ & 64$\to$emb\_dim \\
\hline
Output & Embedding vector & $(B,\text{emb\_dim})$ & Final representation \\
\hline
\end{tabular}
\end{table}

\begin{figure}[htbp]
    \centering
    \includegraphics[width=0.6\linewidth]{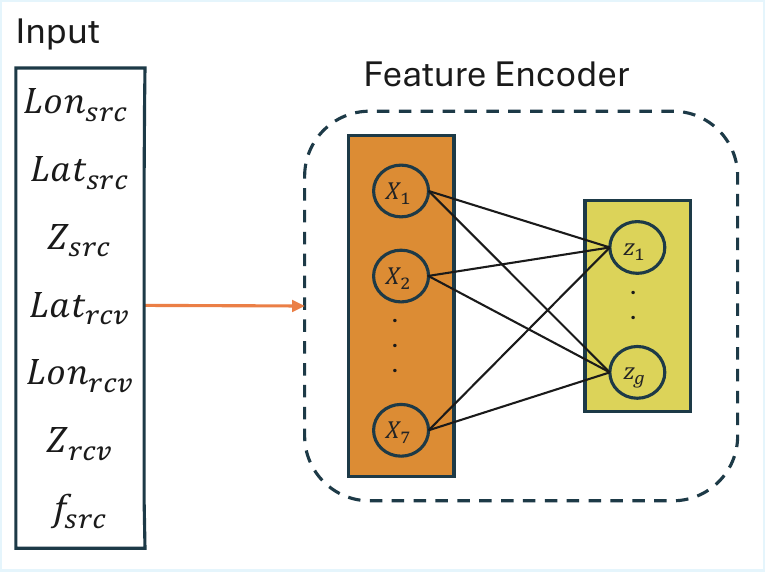}
    \caption{Architecture of the feature encoder. The input vector $\mathbf{x}_g=[\mathbf{X_s},\mathbf{X_r},f]$ is processed through a stack of fully connected layers with batch normalization and GELU activations to produce the latent embedding $\mathbf{z}_g$, which is concatenated with the bathymetry embedding $\mathbf{z}_\Omega$ to form $\mathbf{z}_{\text{lat}}$.}
    \label{fig:feature_encoder_arch}
\end{figure}

\paragraph*{Feature Encoder}  
Geometric configuration also influences transmission loss through source–receiver separation, depth dependence, and frequency content. To capture these effects, the input vector consists of the source coordinates $\mathbf{X_s}=(\text{src\_lat},\text{src\_lon},\text{src\_depth})$, the receiver coordinates $\mathbf{X_r}=(\text{rcv\_lat},\text{rcv\_lon},\text{rcv\_depth})$, and the source frequency $f$. Collectively, the input is :
\[
\mathbf{x}_g = [\mathbf{X_s}, \mathbf{X_r}, f] \in \mathbb{R}^{B \times d_g}, \qquad d_g=7,
\]
where $B$ is the batch size. The encoder is implemented as a multi-layer perceptron (MLP) that maps the input $\mathbf{x}_g$ to a latent embedding $\mathbf{z}_g \in \mathbb{R}^{B \times d_z}$. At an abstract level, the mapping is expressed as : 
\begin{equation}
    \mathbf{z}_g
    = f_{\text{MLP}}\!\big([\mathbf{X_s}, \mathbf{X_r}, f]\big),
    \qquad
    \mathbf{z}_g \in \mathbb{R}^{B \times d_z},
\end{equation}
where $f_{\text{MLP}}$ denotes a composition of affine transformations, batch normalization, and GELU activations. The architecture of the feature encoder is summarized in Table~\ref{tab:feature_encoder}, and the corresponding schematic is shown in Fig.~\ref{fig:feature_encoder_arch}. 
The geometry-derived feature embedding $\mathbf{z}_g$ is concatenated with the bathymetry embedding $\mathbf{z}_\Omega$ to form a combined latent vector:
\begin{equation}
    \mathbf{z}_{\text{lat}} = [\mathbf{z}_g, \mathbf{z}_\Omega],
    \qquad \mathbf{z}_{\text{lat}} \in \mathbb{R}^{B \times (d_z + d_\Omega)}.
\end{equation}
Subsequently, this vector is passed through an additional fully connected block to produce the final latent characteristic $\mathbf{z}$, which serves as input to the residual layer of SVGP. This joint representation encodes both environmental and geometric dependencies of transmission loss.

\begin{table}[htbp]
\centering
\caption{Feature encoder (MLPFeatureExtractor) architecture. Input dimension $d_g{=}7$, output dimension $d_z{=}\texttt{latent\_dim}$ (default $64$). BN = BatchNorm1d.}
\label{tab:feature_encoder}
\begin{tabular}{llll}
\hline
\textbf{Stage} & \textbf{Layer} & \textbf{Output Shape} & \textbf{Details} \\
\hline
Input & Input features & $(B,\,7)$ & $d_g{=}7$ \\
\hline
1 & Linear $\to$ BN $\to$ GELU & $(B,\,32)$ & Linear$(7,32)$, BN$(32)$ \\
2 & Linear $\to$ BN $\to$ GELU & $(B,\,64)$ & Linear$(32,64)$, BN$(64)$ \\
3 & Linear $\to$ BN $\to$ GELU & $(B,\,128)$ & Linear$(64,128)$, BN$(128)$ \\
4 & Linear $\to$ BN $\to$ GELU & $(B,\,256)$ & Linear$(128,256)$, BN$(256)$ \\
5 & Linear $\to$ BN $\to$ GELU & $(B,\,128)$ & Linear$(256,128)$, BN$(128)$ \\
6 & Linear $\to$ BN $\to$ GELU & $(B,\,64)$  & Linear$(128,64)$, BN$(64)$ \\
7 & Linear (output)            & $(B,\,d_z)$ & Linear$(64,d_z)$; no BN/GELU \\
\hline
Output & Feature embedding $\mathbf{z}_g$ & $(B,\,d_z)$ & $d_z{=}\texttt{latent\_dim}$ (default $64$) \\
\hline
\end{tabular}
\end{table}

\subsection{SVGP Residual Head \label{sec:res_head}}
The residual term accounts for unresolved variability in transmission loss beyond spreading and absorption. To capture these effects with calibrated uncertainty at scale, we employ a sparse variational Gaussian process. Let $\mathbf{z}\in\mathbb{R}^{d_z}$ denote the latent feature vector obtained from the neural encoders. A latent Gaussian process $f(\cdot)$ is placed over this space with prior  
\begin{equation}
    f(\cdot) \sim \mathcal{GP}\big(0, k_\phi(\cdot, \cdot)\big),
\end{equation}
where $k_\phi$ is a covariance kernel parameterized by hyperparameters $\phi$. Observations of the residual are assumed to follow a Gaussian likelihood,  
\begin{equation}
    r_i \mid f(\mathbf{z}_i) \sim \mathcal{N}\!\big(f(\mathbf{z}_i), \sigma_n^2\big), \qquad i=1,\dots,N,
\end{equation}
with $\sigma_n^2$ representing the noise variance.  
To capture both short-range variability and multi-scale correlations in the residual, the covariance function is defined as a product kernel,  
\begin{equation}
    k_\phi(\mathbf{z}, \mathbf{z}') \;=\; \sigma_f^2 \, k_{\text{Mat}\,1/2}(\mathbf{z}, \mathbf{z}') \, k_{\text{RQ}}(\mathbf{z}, \mathbf{z}'),
\end{equation}
where $k_{\text{Mat}\,1/2}$ is the Matérn kernel with smoothness parameter $\nu=\tfrac{1}{2}$ (exponential kernel), and $k_{\text{RQ}}$ is the rational quadratic kernel with scale-mixing parameter $\alpha>0$. Explicitly,  
\begin{align}
    k_{\text{Mat}\,1/2}(\mathbf{z}, \mathbf{z}') &= \exp\!\left(- \sqrt{\sum_{d=1}^{d_z}\frac{(z_d - z'_d)^2}{\ell_{\text{Mat},d}^2}}\right), \\
    k_{\text{RQ}}(\mathbf{z}, \mathbf{z}') &= \left(1 + \frac{1}{2\alpha}\sum_{d=1}^{d_z}\frac{(z_d - z'_d)^2}{\ell_{\text{RQ},d}^2}\right)^{-\alpha},
\end{align}
with $\{\ell_{\text{Mat},d}\}$ and $\{\ell_{\text{RQ},d}\}$ denoting automatic relevance determination (ARD) length scales for each latent dimension.  
For computational scalability, the SVGP introduces $M \ll N$ inducing points $\mathbf{Z}=\{\mathbf{z}_m\}_{m=1}^M$ with corresponding inducing variables $\mathbf{u}=f(\mathbf{Z})$. The prior distribution over the inducing variables is  
\begin{equation}
    p(\mathbf{u}) = \mathcal{N}\!\big(\mathbf{0}, \mathbf{K}_{ZZ}\big), \qquad [\mathbf{K}_{ZZ}]_{mn}=k_\phi(\mathbf{z}_m,\mathbf{z}_n).
\end{equation}
A variational Gaussian distribution  
\begin{equation}
    q(\mathbf{u}) = \mathcal{N}(\mathbf{m}, \mathbf{S}),
\end{equation}
is introduced, yielding the approximate posterior  
\begin{equation}
    q(f) = \int p(f \mid \mathbf{u}) q(\mathbf{u}) \, d\mathbf{u}.
\end{equation}

The predictive distribution of the residual at a new test input $\mathbf{z}_\star$ is then Gaussian with mean and variance  
\begin{align}
    \mu_{\text{GP}}(\mathbf{z}_\star) &= \mathbf{k}_{\star Z}\mathbf{K}_{ZZ}^{-1}\mathbf{m}, \\
    \sigma^2_{\text{GP}}(\mathbf{z}_\star) &= k_\phi(\mathbf{z}_\star, \mathbf{z}_\star) - \mathbf{k}_{\star Z}\mathbf{K}_{ZZ}^{-1}\mathbf{k}_{Z\star} + \mathbf{k}_{\star Z}\mathbf{K}_{ZZ}^{-1}\mathbf{S}\,\mathbf{K}_{ZZ}^{-1}\mathbf{k}_{Z\star},
\end{align}
where $\mathbf{k}_{\star Z}=[k_\phi(\mathbf{z}_\star, \mathbf{z}_1),\dots,k_\phi(\mathbf{z}_\star, \mathbf{z}_M)]$. 
The final transmission loss prediction combines the physics-informed mean with the GP residual mean,  
\begin{equation}
    \widehat{\mathrm{TL}} = \mathrm{TL}_{\text{phys}}(R,f) + \mu_{\text{GP}}(\mathbf{z}),
\end{equation}
and its predictive uncertainty is given by  
$\mathrm{Var}\!\left[\widehat{\mathrm{TL}}\right] = \sigma^2_{\text{GP}}(\mathbf{z})$.
The SVGP is trained by maximizing the evidence lower bound. For a mini-batch $\mathcal{B}$ of size $B$, the objective takes the form  
\begin{equation}
    \mathcal{L}_{\text{ELBO}}(\mathcal{B}) = -\frac{N}{B}\sum_{i\in\mathcal{B}} \mathbb{E}_{q(f)}\!\left[\log \mathcal{N}\!\big(r_i \mid f(\mathbf{z}_i), \sigma_n^2\big)\right] + \mathrm{KL}\!\big(q(\mathbf{u}) \,\|\, p(\mathbf{u})\big),
\end{equation}
which balances the fidelity of the residual fit with a complexity penalty on the variational approximation.  
To ensure conservative predictions for operational use, the ELBO is augmented with a one-sided penalty that discourages overestimation of transmission loss:  
\begin{equation}
    \mathcal{L}_{\text{pen}} = \frac{1}{B}\sum_{i\in\mathcal{B}} \Big[\max\!\big(0, \,  \widehat{\mathrm{TL}}^{(i)} - \mathrm{TL}_{\text{max}}  \big)\Big]^2.
\end{equation}
The overall loss is therefore given by: 
\begin{equation}
    \mathcal{L} = \mathcal{L}_{\text{ELBO}} + \lambda \, \mathcal{L}_{\text{pen}},
    \label{eqn:tot_loss}
\end{equation}
with $\lambda$ a weighting hyper-parameter. This loss function ensures statistically calibrated and operationally conservative predictions, making the surrogate suitable for real-time digital twin deployment.

In contrast to purely data-driven surrogates such as convolutional neural networks or deep Gaussian process variants (e.g., DSPP, SVDKL), the proposed framework integrates analytic physics-based structure with scalable probabilistic learning. By embedding geometric spreading and absorption in the mean function, the surrogate retains the interpretability and extrapolation capability. The neural encoders provide compact latent representations of environmental and geometric variability, while the SVGP residual supplies calibrated uncertainty and multiscale correlation structure. This combination yields a digital twin that is both computationally efficient and operationally reliable in complex coastal acoustic environments.


\section{Data Generation and Training}
\label{sec:training}
\begin{figure}[htbp]
\centering
\begin{subfigure}[t]{0.48\linewidth}
\centering
\includegraphics[height=6cm]{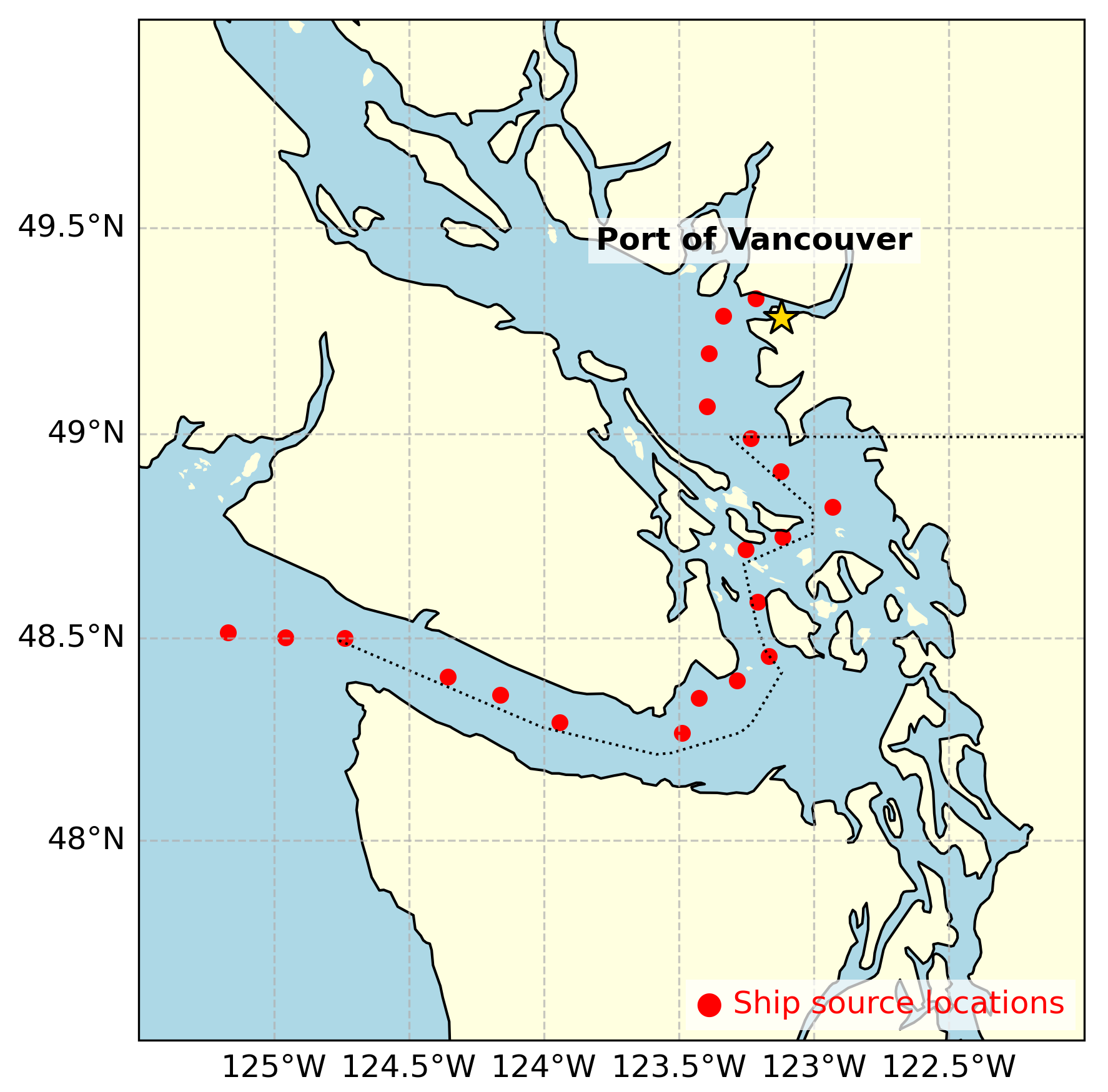}
\caption{Locations of the 20 ship sources (red dots) along the primary shipping route to the Port of Vancouver (yellow star).}
\label{fig:ship_source_locations_map}
\end{subfigure}
\hfill
\begin{subfigure}[t]{0.48\linewidth}
\centering
\includegraphics[height=6cm]{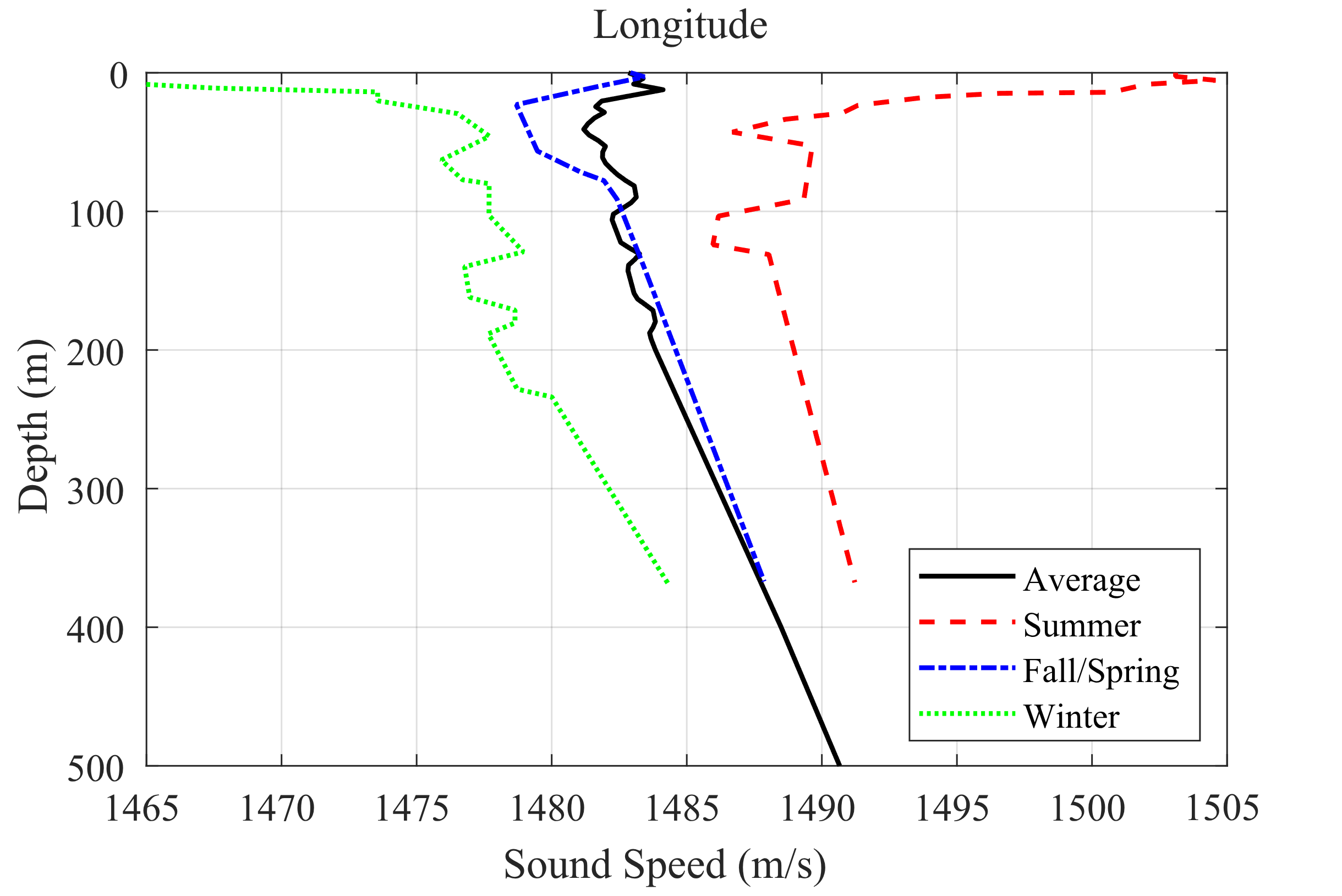}
\caption{Empirical sound speed profiles: average (black), summer (red), fall/spring (blue), and winter (green).}
\label{fig:sound_speed_profile}
\end{subfigure}
\caption{Ship source locations and sound speed profiles used in the simulations.}
\label{fig:data_generation}
\end{figure}

Training and evaluation data are generated using Gaussian beam-based ray tracing with the Bellhop3D solver \cite{porter1987gaussian, porter_bellhop_nodate}. The simulations are parameterized by source coordinates, receiver coordinates, and source frequency, and can be conducted for arbitrary ocean environments given appropriate bathymetry and sound-speed profiles. As a demonstration, we consider the Vancouver coastal region, using seasonally averaged sound-speed profiles derived from empirical measurements (Fig.~\ref{fig:sound_speed_profile}). A total of 20 ship source locations are sampled along the primary shipping route to the Port of Vancouver (Fig.~\ref{fig:ship_source_locations_map}). For each source, receivers are uniformly distributed within a 100 km horizontal radius and to depths of up to 110 m, yielding a spatially resolved acoustic field. The simulations are conducted across 30 frequency components corresponding to one-third octave bands ranging from 12.5 Hz to 8 kHz, thereby capturing the broadband character of ship-generated noise. Transmission loss is computed from the simulated pressure fields according to Eq.~\eqref{eq:TL_comp}, where $p(\mathbf{X}_{rcv},\omega_{src})$ denotes the acoustic pressure at receiver location $\mathbf{X}_{rcv}$ and frequency $\omega_{src}$, and $p_{\text{ref}}$ is the source amplitude at a reference distance. This setup produces a dataset of more than 30 million source–receiver pairs, providing the scale required for training the probabilistic surrogate.

The Bellhop solver approximates solutions of the Helmholtz equation using Gaussian beam ray tracing, accounting for range-dependent sound speed variability, geometric spreading, and boundary interactions at the surface and seabed boundary interactions. For each source, rays are launched across a full range of take-off angles, with the solver adaptively determining the number of beams and integration step size. This procedure yields snapshots of TL distributions for each source and frequency combination, forming a comprehensive dataset for model training. Of the 20 source locations, 75\% of the source–receiver–frequency samples are used for training, 15\% for validation, and 10\% are held out for testing. In total, more than 30 million data points are generated, covering source–receiver pairs distributed along the shipping channel from the Pacific Ocean to the Port of Vancouver. This dataset provides the input–output pairs required to train the probabilistic digital twin, linking source–receiver geometry, frequency, and bathymetric information to transmission loss predictions with quantified uncertainty.

\begin{figure}[htbp]
    \centering
    \includegraphics[width=0.95\linewidth]{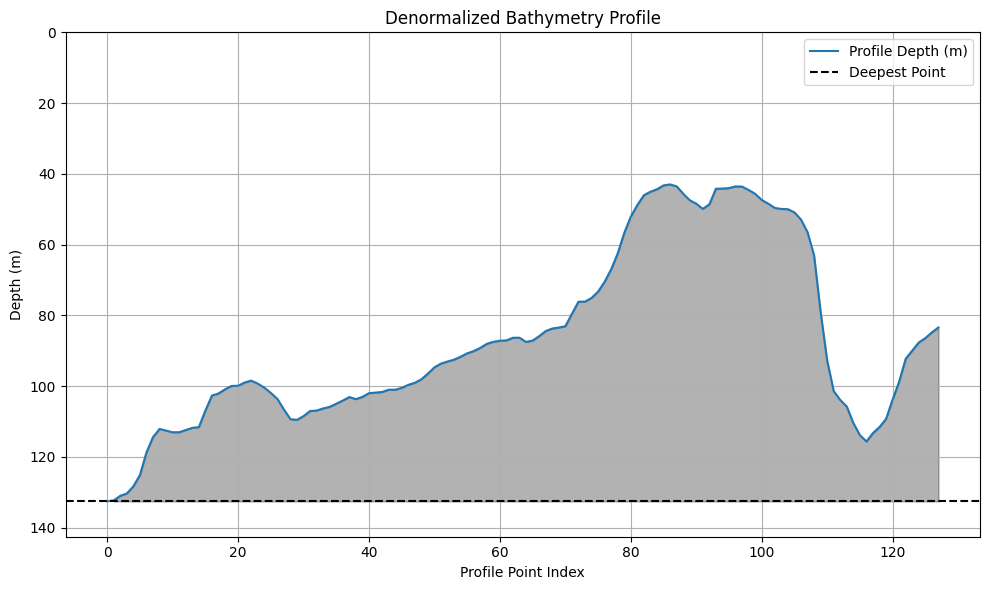}
    \caption{Example of a denormalized bathymetry profile sampled along the source–receiver path. The profile is discretized into 128 uniformly spaced points, consistent with the representation $\Omega \in \mathbb{R}^{128}$.}
    \label{fig:bathy_profile}
\end{figure}

The model input is represented by an eight-dimensional feature vector,
\begin{equation}
\mathbf{x} = \bigl(\text{src}_{\text{lat}},\ \text{src}_{\text{lon}},\ \text{src}_{\text{depth}},\ 
\text{rcv}_{\text{lat}},\ \text{rcv}_{\text{lon}},\ \text{rcv}_{\text{depth}}, \Omega, \text{frequency}\bigr),
\end{equation}
where $(\text{src\_lat},\ \text{src\_lon},\ \text{src\_depth})$ denote the latitude, longitude, and depth of the acoustic source, and $(\text{rcv\_lat},\ \text{rcv\_lon},\ \text{rcv\_depth})$ define the corresponding receiver location. The bathymetric profile is represented by $\Omega \in \mathbb{R}^{128}$, a vector obtained by sampling the seafloor elevation along the straight line connecting source and receiver at 128 uniformly spaced points using the GEBCO dataset \cite{GEBCO_2024}. An example of such a denormalized bathymetry profile is illustrated in Fig.~\ref{fig:bathy_profile}, where the seabed elevation is sampled along the propagation path. The final input feature corresponds to the source frequency. All variables are normalized to ensure consistent scaling prior to training. Latitudes and longitudes are linearly mapped to the unit interval according to:
\begin{equation}
\text{lat}_{\text{norm}} = \frac{\text{lat}+90}{180},
\qquad
\text{lon}_{\text{norm}} = \frac{\text{lon}+180}{360},
\end{equation}
while depths, bathymetry and frequencies are min--max scaled according to their observed ranges given by:
\begin{equation}
x_{\text{norm}} = \frac{x - x_{\min}}{x_{\max} - x_{\min}},
\qquad x \in \{\text{src\_depth},\ \text{rcv\_depth}, \Omega, \text{frequency}\}.
\end{equation}

This transformation ensures that all features lie in $[0,1]$, preventing scale imbalances during optimization. 
The response variable considered in this study is the transmission loss, expressed in decibels relative to a reference pressure at one meter from the source. In order to mitigate the impact of numerical artifacts arising from ray tracing, particularly in regions of shadow zones or near caustics, TL values exceeding 200 dB are clipped at this threshold \cite{deo2025predicting, mallik2022predicting}. This procedure avoids the introduction of extreme outliers that could adversely influence the training process while remaining consistent with physically plausible ranges of underwater acoustic propagation. This choice preserves physical interpretability and was empirically found to enhance stability in the variational training procedure.

\subsection{Training}
The model is trained using a stochastic variational framework \cite{hoang2015unifying} in which the residual between the physics-informed mean and the observed transmission loss is modeled by a sparse variational Gaussian process. 
Let $(\mathbf{x}_i,y_i)$ denotes the $i$-th training example (input features, and transmission loss), and let
$\mathbf{z}_i = f_{\text{enc}}(\mathbf{x}_i) \in \mathbb{R}^{d_z}$ be the latent feature produced by the neural encoders as given in the section \ref{sec:neural_enc}.
The physics-informed mean is denoted $\mathrm{TL}_{\text{phys}}(\mathbf{x}_i)$, and the residual target is :
\begin{equation}
    r_i \;=\; y_i - \mathrm{TL}_{\text{phys}}(\mathbf{x}_i).
\end{equation}
The SVGP residual head (Sec.~\ref{sec:res_head}) yields a latent GP posterior $q(f)$ with inducing variables and variational parameters.
For a mini-batch $\mathcal{B}$ of size $B$, the stochastic objective is the negative ELBO augmented with a hinge penalty given by Eq.~\ref{eqn:tot_loss} with penalty weight $\lambda=10.0$ in our experiments. The hinge term discourages predictions that exceed a physically conservative upper bound $\mathrm{TL}_{\max}=200~\mathrm{dB}$.

Training continues with the AdamW optimizer \cite{loshchilov2017decoupled}, which combines the Adam \cite{kingma2017adam} adaptive moment estimation scheme with decoupled weight decay regularization. In our implementation, the initial learning rate is set to $\eta_{0}=10^{-3}$ with weight decay $\beta=10^{-2}$, and $\beta_{1}=0.9$, $\beta_{2}=0.999$ for the exponential moving averages of the first and second moments. Training is performed on mini-batches that are reshuffled at every iteration to reduce sampling bias. To improve numerical robustness of GP solves, a dynamic Cholesky jitter $\epsilon$ is employed inside the linear algebra routines; On detection of a Cholesky failure, $\epsilon$ is multiplicatively increased and the mini-batch is discarded. To prevent exploding gradients, we apply $\ell_2$-norm clipping,
\begin{equation}
    \|\nabla_{\Theta}\mathcal{L}\|_2 \;\leftarrow\; \min\!\Big(\|\nabla_{\Theta}\mathcal{L}\|_2,\; \gamma\Big),
    \qquad \gamma = 1.0,
\end{equation}
where $\Theta$ denotes all trainable parameters (encoders, GP hyperparameters, inducing locations, and variational parameters). The learning rate is adapted using a cosine annealing schedule \cite{loshchilov2016sgdr}. Specifically, over the course of $T_{\max}$ epochs, the learning rate at epoch $t$ is updated as :
\begin{equation}
    \eta_t \;=\; \eta_{\min} + \tfrac{1}{2}\,(\eta_{\max}-\eta_{\min})\Big[1+\cos\!\Big(\pi \tfrac{t}{T_{\max}}\Big)\Big],
\end{equation}
where $t$ is the current epoch index, $\eta_{\max}=\eta_{0}$ is the initial learning rate, and $\eta_{\min}=10^{-6}$ is the minimum learning rate. Here, $T_{\max}$ denotes the total number of training epochs such that the learning rate decays smoothly from $\eta_{\max}$ to $\eta_{\min}$ following a single cosine cycle.
At the end of each epoch, we evaluate a holdout validation set using: (i) negative ELBO (without hinge penalty) as the selection criterion,
(ii) mean squared error (MSE) between $\widehat{\mathrm{TL}}$ and $y$, and
(iii) root mean square percentage error (RMSPE), computed as:
\begin{equation}
    \mathrm{RMSPE}~[\%] \;=\; 100 \times
    \sqrt{\frac{1}{|\mathcal{V}|}\sum_{i\in\mathcal{V}}
          \left(\frac{\widehat{\mathrm{TL}}_i - y_i}{y_i + \varepsilon}\right)^2},
    \quad \varepsilon = 10^{-6}.
\end{equation}
To align validation with operational constraints, the validation mean prediction is clamped above by $\mathrm{TL}_{\max}$:
$\widehat{\mathrm{TL}}_i \leftarrow \min(\widehat{\mathrm{TL}}_i,\mathrm{TL}_{\max})$.
We adopt early stopping based on the validation ELBO with patience $\Pi=30$ epochs. 
Let $E^\star$ be the epoch with the lowest validation loss observed so far. If no improvement beyond a tolerance of $10^{-6}$ is observed for $\Pi$ consecutive epochs, training stops and the best model state (at $E^\star$) is restored. The overall training procedure is summarized in Algorithm~\ref{alg:training_loop}. 
The complete model is implemented on a workstation equipped with an Intel Xeon~W5 CPU, an NVIDIA RTX~6000 Ada GPU, and 128\,GB of system memory. The framework is developed in \texttt{PyTorch}~\cite{paszke2019pytorch}, version~2.5.1, compiled with CUDA~12.4 support. Gaussian process components are implemented using \texttt{GPyTorch}~\cite{gardner2018gpytorch}, version~1.13.

\begin{algorithm}[H]
\caption{\textbf{Training Algorithm: Mini-batch SVGP with Hinge Penalty}}
\label{alg:training_loop}
\begin{algorithmic}[1]
\STATE \textbf{Input:} Training set $\mathcal{D}=\{(\mathbf{x}_i,\Omega_i,y_i)\}_{i=1}^N$, batch size $B$, penalty weight $\lambda$, max TL $\mathrm{TL}_{\max}$, gradient clip $\gamma$, patience $\Pi$
\STATE \textbf{Output:} Parameters $\Theta$ (encoders, GP hyperparameters, inducing locations, variational mean/covariance, likelihood)
\STATE Initialize $\Theta$ (including inducing inputs and variational parameters); set $L_{\text{best}}\gets\infty$, $p\gets 0$
\FOR{epoch $= 1$ to $N_{\mathrm{epochs}}$}
    \STATE Shuffle $\mathcal{D}$ and split into mini-batches $\{\mathcal{B}_j\}$ of size $B$
    \FOR{each mini-batch $\mathcal{B}$}
        \STATE \textbf{Forward:} For $(\mathbf{x}_i,\Omega_i,y_i)\in\mathcal{B}$, compute $\mathbf{z}_i \leftarrow f_{\text{enc}}(\mathbf{x}_i,\Omega_i)$ and $r_i \leftarrow y_i - \mathrm{TL}_{\text{phys}}(\mathbf{x}_i)$
        \STATE Evaluate $\mathcal{L}_{\text{ELBO}}(\mathcal{B}) \leftarrow -\frac{N}{B}\sum_{i\in\mathcal{B}}\mathbb{E}_{q(f)}\!\left[\log \mathcal{N}\!\big(r_i \mid f(\mathbf{z}_i),\sigma_n^2\big)\right] + \mathrm{KL}\!\big(q(\mathbf{u})\|p(\mathbf{u})\big)$
        \STATE Form predictions $\widehat{\mathrm{TL}}_i \leftarrow \mathrm{TL}_{\text{phys}}(\mathbf{x}_i)+\mu_{\text{GP}}(\mathbf{z}_i)$
        \STATE Compute hinge penalty $\mathcal{L}_{\text{hinge}}(\mathcal{B}) \leftarrow \frac{1}{B}\sum_{i\in\mathcal{B}}\big[\max(0,\widehat{\mathrm{TL}}_i-\mathrm{TL}_{\max})\big]^2$
        \STATE Total loss $\mathcal{L}(\mathcal{B}) \leftarrow \mathcal{L}_{\text{ELBO}}(\mathcal{B}) + \lambda\,\mathcal{L}_{\text{hinge}}(\mathcal{B})$
        \STATE \textbf{Backward/Update:} $\hat{\mathbf{g}} \leftarrow \nabla_{\Theta}\mathcal{L}(\mathcal{B})$; clip $\|\hat{\mathbf{g}}\|_2 \leftarrow \min(\|\hat{\mathbf{g}}\|_2,\gamma)$; update $\Theta$ (e.g., AdamW); zero grads
    \ENDFOR
    \STATE \textbf{Validation:} Compute $L_{\text{val}}$ (with predictions clamped at $\mathrm{TL}_{\max}$), MSE, and RMSPE; step learning-rate scheduler
    \IF{$L_{\text{val}} < L_{\text{best}} - 10^{-6}$}
        \STATE Save checkpoint; $L_{\text{best}}\gets L_{\text{val}}$; $p\gets 0$
    \ELSE
        \STATE $p \gets p + 1$
    \ENDIF
    \IF{$p \ge \Pi$} \STATE \textbf{break} \ENDIF
\ENDFOR
\STATE \textbf{return} $\Theta$
\end{algorithmic}
\end{algorithm}
\section{Results and Discussion}
\label{sec:results}
In this section, we assess the predictive performance of the proposed probabilistic digital twin for underwater acoustics across large-scale transmission loss datasets. The analysis begins with baseline sparse variational GP formulations, extends to hierarchical GP models, and culminates in the physics-guided probabilistic surrogate. The final architecture, which integrates feature and bathymetry encoders with an SVGP residual component, is tested using datasets generated with the Bellhop3D Gaussian beam solver. As a demonstration, we focus on the Salish Sea region, where three-dimensional simulations capture range-dependent variability along major shipping routes. Finally, we present operational case studies, including vessel transit, dynamic speed optimization, and sensor assimilation, to illustrate how the framework supports adaptive, uncertainty-aware digital twins for underwater noise management.

\subsection{Baseline SVGP with Zero-Mean and Stationary Kernel}
\label{sec:baseline_svgp}
\begin{figure}[htbp]
    \centering
    \makebox[\textwidth][c]{%
        \includegraphics[width=1.3\linewidth]{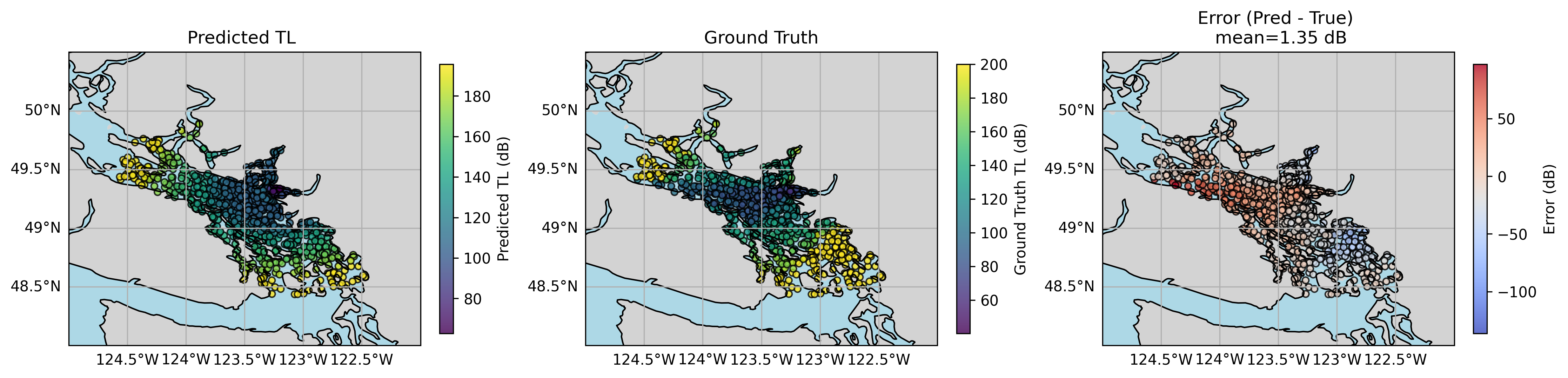}%
    }
    \caption{Baseline SVGP with zero mean prior and Matérn--RBF kernel. 
    Left: predicted TL. Middle: ground-truth TL. Right: error (prediction minus truth); 
    the mean signed error ($1.35$\,dB).}
    \label{fig:baseline_svgp_map}
\end{figure}

We first evaluate a sparse variational Gaussian process (SVGP) with a zero mean function and a Matérn--RBF kernel. Figure~\ref{fig:baseline_svgp_map} compares predicted transmission loss at depth of 40 m and source frequency of 400 Hz, the ground truth, and the corresponding error field. The mean signed error is $1.35$\,dB, which indicates that the overall bias is small. However, the error field shows a wide spread across the spatial domain, the standard deviation of residual is 28.56. This behavior can be explained as follows. (i) The zero-mean prior forces the kernel to represent both the dominant physical trend (geometric spreading and absorption) and the residual fluctuations, which reduces the effective flexibility of the kernel. (ii) The Matérn--RBF kernel is stationary, so it imposes a fixed correlation structure across the domain. This assumption does not hold in regions with range-dependent sound speed or variable bathymetry, which leads to large residuals.  (iii) The variational approximation with a finite number of inducing points restricts the ability of the model to resolve fine-scale structure in regions where the receiver distribution is dense.  
As a result, the model achieves a small mean error, but exhibits a large variance in prediction error. This motivates the addition of physics-informed mean functions and nonstationary latent representations through learned encoders, which are explored in the following sections.  

\subsection{Hierarchical GP Model: Deep Sigma Point Process}
\label{sec:dspp_results}

\begin{figure}[htbp]
    \centering
    \makebox[\textwidth][c]{%
        \includegraphics[width=1.3\linewidth]{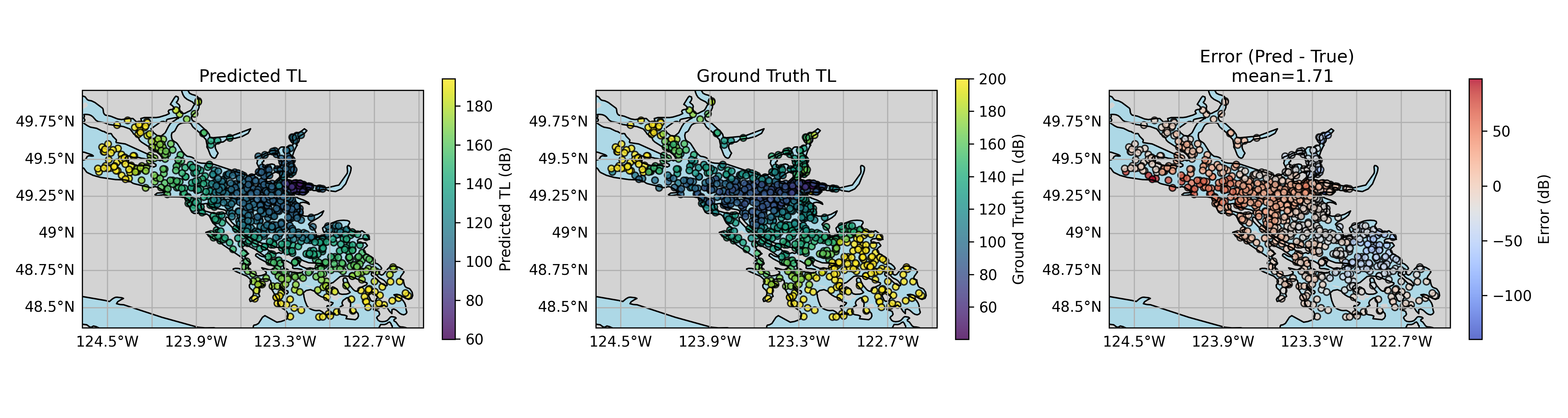}%
    }
    \caption{Baseline SVGP (zero mean, Matérn--RBF kernel) over the evaluation domain. 
    Left: predicted TL. Middle: ground-truth TL. Right: error (prediction minus truth); 
    the mean signed error ($1.35$\,dB).}
    \label{fig:dspp_map}
\end{figure}

We next evaluate the deep sigma point process , a hierarchical GP model designed to capture non-Gaussian latent functions through sigma-point sampling. Figure~\ref{fig:dspp_map} shows the predicted transmission loss at depth of 40 m and source frequency of 400 Hz, ground truth, and error distribution for the DSPP model. The mean signed error is $1.71$\,dB, which is comparable to the baseline SVGP. However, the error spread is reduced in several coastal regions where bathymetric variability is high, the standard deviation of residual is 21.30. This improvement arises because the DSPP introduces a more flexible latent representation than a stationary kernel, allowing it to better resolve nonstationary correlations in range-dependent environments.

Despite these improvements, the DSPP also exhibits residual errors in areas with sharp bathymetric transitions, as indicated by localized regions of over- and under-prediction. The higher computational cost of training DSPP compared to SVGP is a trade-off for the gain in flexibility. These results indicate that hierarchical GP models such as DSPP can partially overcome the limitations of stationary kernels by capturing multi-scale variability, but further refinements are required to consistently reduce error across the domain.

\subsection{Proposed Physics-Guided Encoder--SVGP Model}
\label{sec:final_model}

We now evaluate the proposed architecture that integrates a physics-informed mean function with learned bathymetry and feature encoders, followed by an SVGP residual head. The physics-informed mean captures the dominant effects of geometric spreading and frequency-dependent absorption, while the bathymetry encoder processes range-dependent seabed variability and the feature encoder represents source--receiver geometry and frequency. The concatenated embeddings provide a nonstationary latent representation, enabling the SVGP residual layer to correct the remaining mismatch with calibrated uncertainty estimates.

\begin{figure}[htbp]
    \centering
    \makebox[\textwidth][c]{%
        \includegraphics[width=1.3\linewidth]{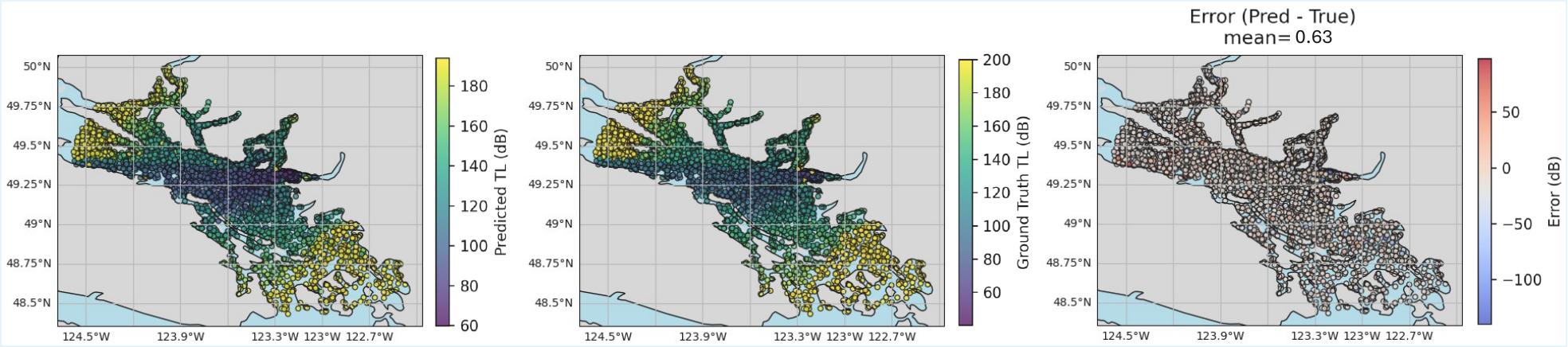}%
    }
    \caption{Results for the proposed encoder--SVGP model. Left: predicted transmission loss (TL). Middle: ground truth TL. Right: prediction error. The model achieves the lowest prediction error and the best-calibrated uncertainty among all tested models.}
    \label{fig:final_fields}
\end{figure}

Figure~\ref{fig:final_fields} shows the predicted transmission loss  at depth of 40 m and source frequency of 400 Hz, the reference ground truth, and the associated prediction error. The proposed framework yields the lowest mean signed error 0.63. Compared to baseline SVGP and hierarchical GP models, the encoder--SVGP model reduces both the overall bias and the spatial spread of residual errors, particularly in regions with range-dependent bathymetry and strong sound-speed gradients. The standard deviation of the residual is 15.20.

\subsection{Ablation and Learning Capacity}
\label{sec:discussion}

To further interpret the performance of the proposed encoder--SVGP model, we conduct an ablation study to assess the role of the physics-informed mean and the learning capacity of the neural encoders. Removing the physics-informed mean and training the SVGP with a zero mean significantly increases both the mean signed error and the variance of the residuals, as the GP must then learn the dominant deterministic trend associated with spherical spreading and absorption. Introducing the physics-informed mean reduces this burden, allowing the GP residual head to focus on range-dependent fluctuations driven by bathymetry and sound-speed variability. This confirms that embedding physics into the mean function enhances data efficiency and improves predictive stability.

The implicit bias introduced by the bathymetry and feature encoders is also critical. These encoders transform the raw source--receiver geometry and seabed profile into a latent representation that enables the GP to capture spatially varying correlation structures. Without these encoders, the SVGP operates in the raw input space with stationary kernels, which limits its ability to resolve localized variability. The improvement observed in the standard deviation of the residuals demonstrates that the combined architecture not only reduces the average error, but also suppresses the spread of prediction errors across the domain.

Table~\ref{tab:ablation_error} summarizes the mean signed error and standard deviation of the residuals for different model variants. The results show that the inclusion of the physics-informed mean and learned encoders systematically improves both accuracy and robustness.
These results demonstrate that the integration of physics priors and implicit bias through encoders yields the most balanced architecture, combining low bias, reduced variance, and calibrated uncertainty.

\begin{table}[htbp]
\centering
\caption{Mean signed error and standard deviation of residuals for different model variants.}
\label{tab:ablation_error}
\begin{tabular}{lcc}
\hline
Model Variant & Mean Signed Error (dB) & Std. Dev. of Residuals (dB) \\
\hline
SVGP (Zero Mean, Matérn--RBF) & 1.35 & 28.56 \\
SVGP (Physics-Informed Mean)  & 1.21 & 24.67 \\
DSPP                         & 1.75 & 21.30 \\
Proposed Encoder--SVGP       & \textbf{0.63} & \textbf{15.20} \\
\hline
\end{tabular}
\end{table}

\subsection{Case Study: Ship Transit Along the Vancouver Shipping Channel}
\label{sec:ship_channel}

To illustrate the operational capability of the proposed probabilistic digital twin, we consider a ship transiting from the open Pacific Ocean toward the Port of Vancouver along the main shipping channel. At three representative source locations along the route, transmission loss (TL) maps are generated and compared. The framework predicts the full three-dimensional TL field within a $100$~km horizontal radius and down to a depth of $110$~m for each source in approximately $300$~ms, whereas the Bellhop 3D solver requires about $4$ minutes for all frequencies in a one third ocatave band from 12.5 Hz to 8 kHz. This corresponds to an acceleration of nearly $800\times$, enabling near real-time prediction.  

\begin{figure}[htbp]
\centering
    \makebox[\textwidth][c]{%
    \centering
    \begin{subfigure}[t]{0.42\linewidth}
        \centering
        \includegraphics[width=\linewidth]{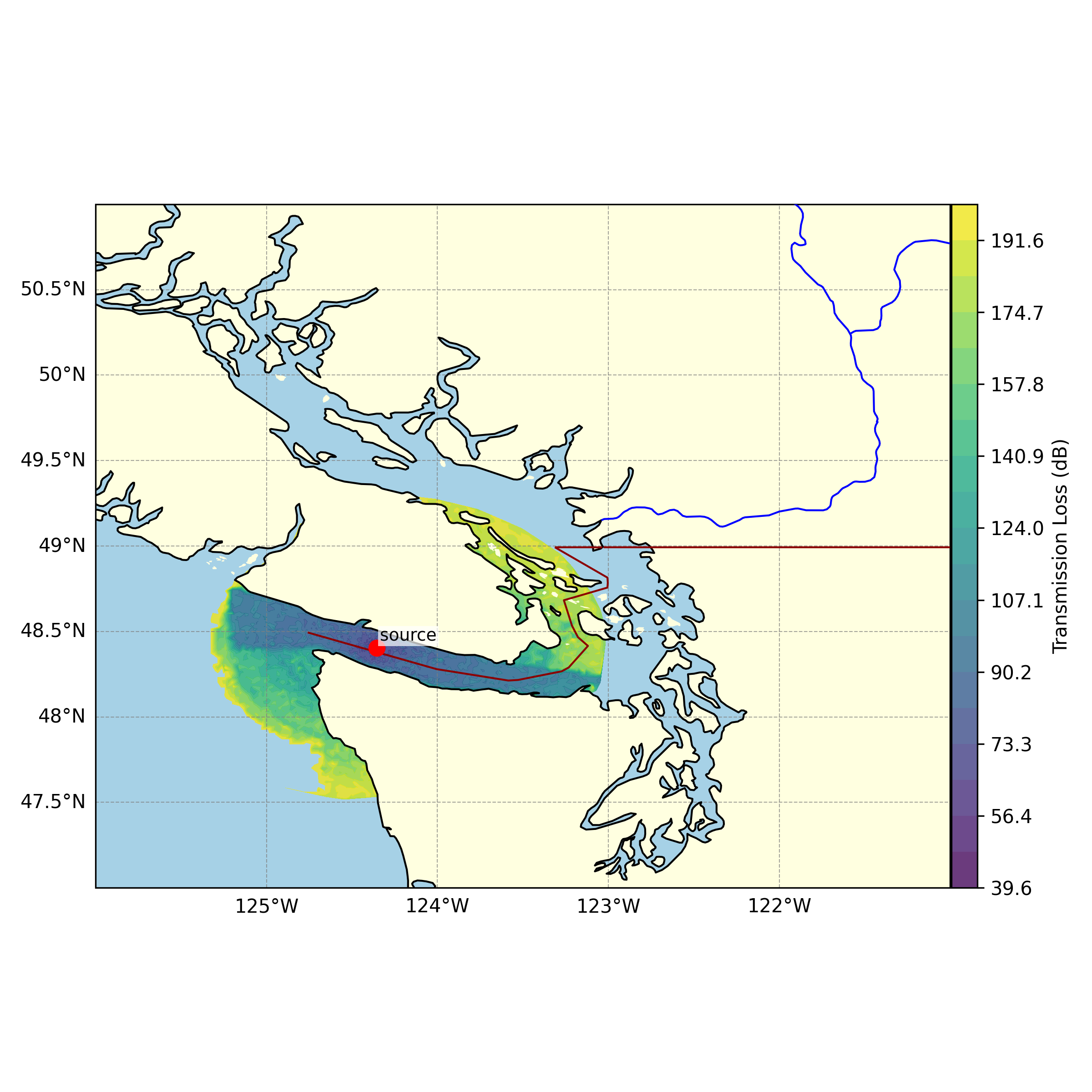}
        \caption{Source at offshore location.}
    \end{subfigure}
    \hfill
    \begin{subfigure}[t]{0.42\linewidth}
        \centering
        \includegraphics[width=\linewidth]{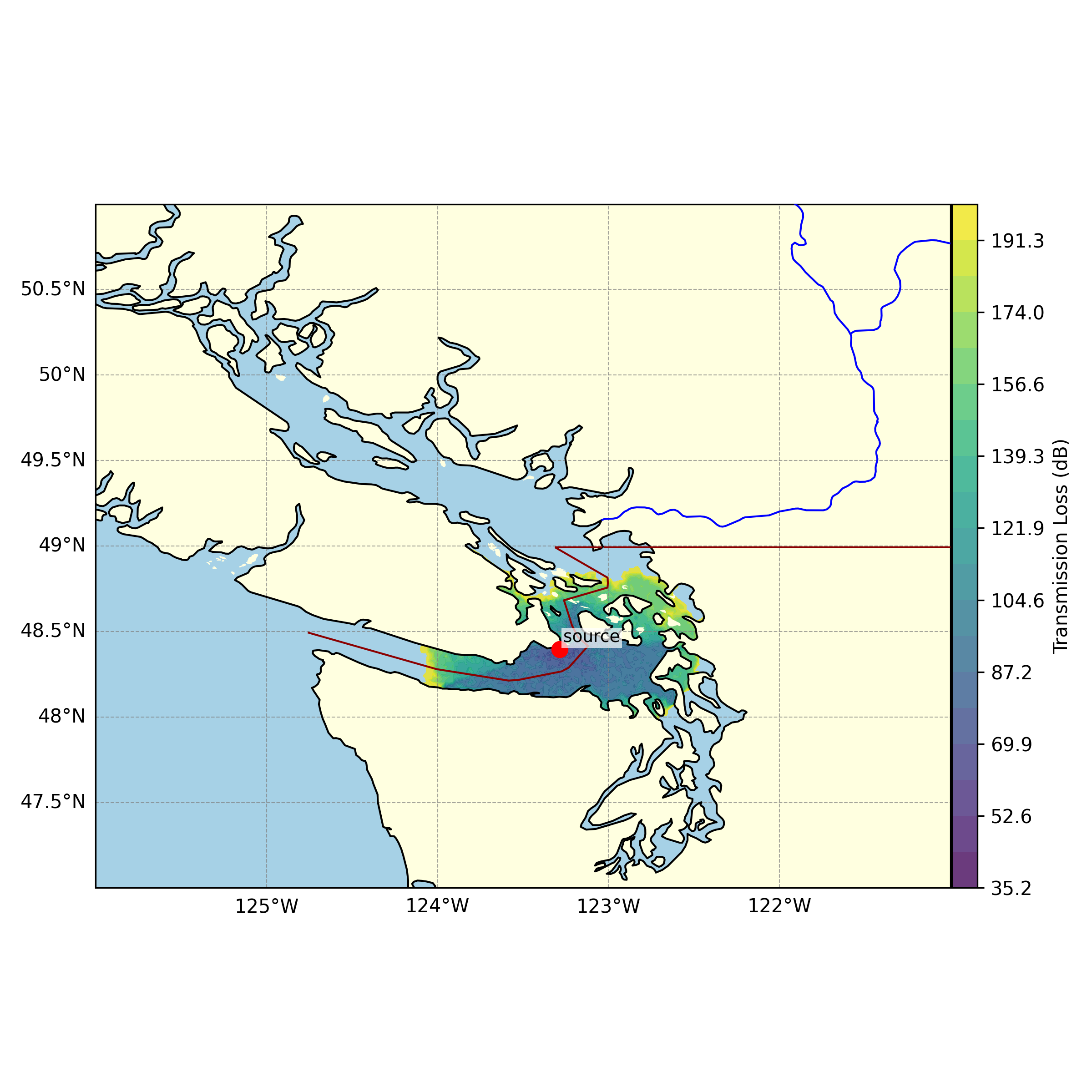}
        \caption{Source at mid-channel.}
    \end{subfigure}
    \hfill
    \begin{subfigure}[t]{0.42\linewidth}
        \centering
        \includegraphics[width=\linewidth]{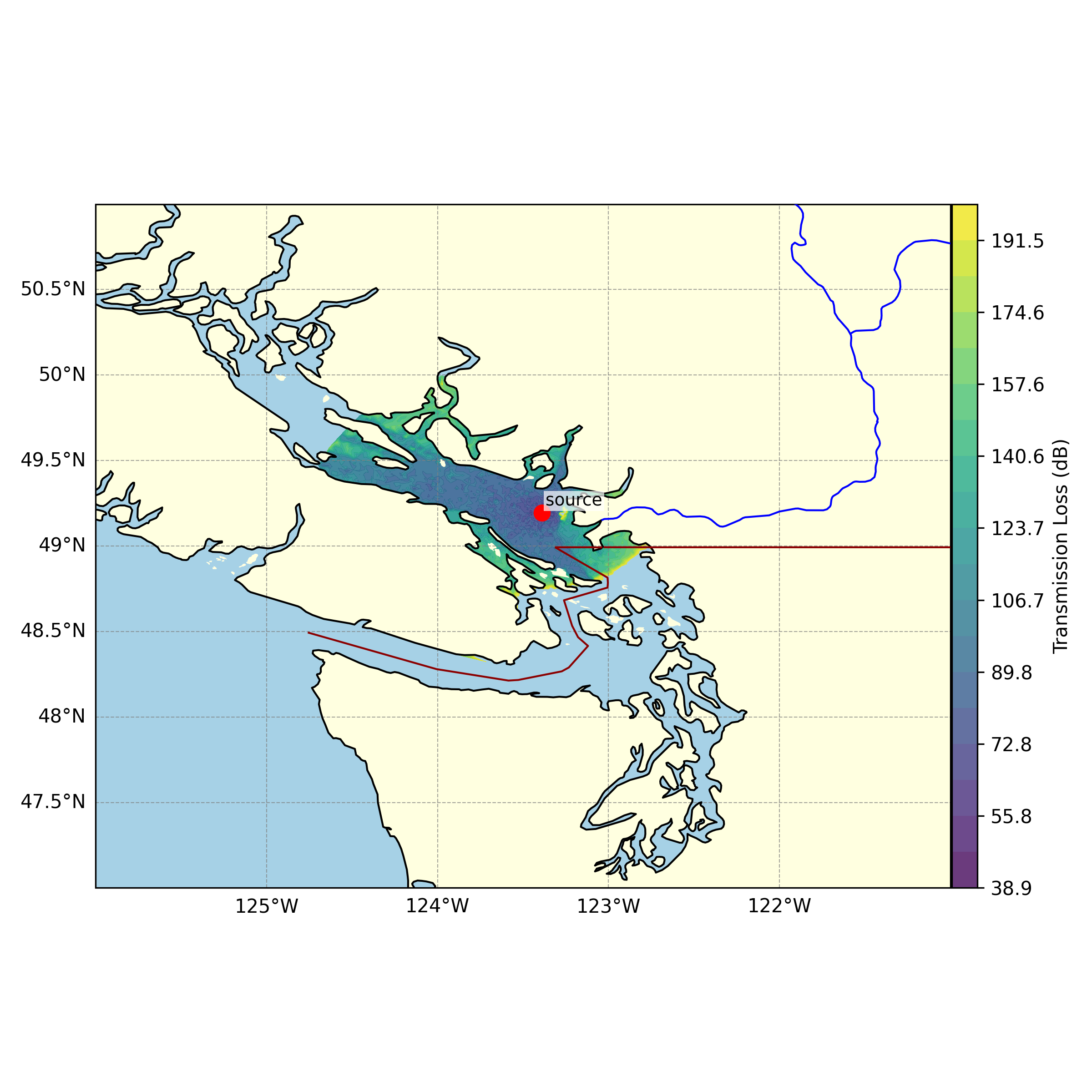}
        \caption{Source near Port of Vancouver.}
    \end{subfigure}
}
    \caption{Predicted transmission loss maps at depth $40$~m for a source frequency of $400$~Hz at three source locations along the shipping channel from the open Pacific Ocean to the Port of Vancouver. The framework produces the full 3D TL field within $300$~ms, achieving an $800\times$ speedup compared to Bellhop 3D.}
    \label{fig:ship_channel_tlmap}
\end{figure}

Figure~\ref{fig:ship_channel_tlmap} shows TL maps at a receiver depth of $40$~m for a source frequency of $400$~Hz at three distinct source positions along the channel. The predicted fields capture the range-dependent variability induced by bathymetry and environmental conditions, while maintaining close agreement with the ground-truth solver. These results highlight the ability of the proposed framework to track evolving acoustic footprints of vessels during transit and to provide fast updates required for operational noise mitigation strategies.

\subsection{Operational Digital Twin: Dynamic speed optimization along a fixed route}
\label{sec:ops_speed_opt}
We next demonstrate how the proposed surrogate framework can be deployed in an operational setting to dynamically adjust vessel speed along a prescribed route with the objective of minimizing cumulative sound exposure at a marine mammal location. By combining near–field source levels predicted using the JOMOPANS–ECHO model \cite{macgillivray2021reference}, transmission loss estimated by the surrogate, and the resulting received levels at the receiver location, the framework enables per–leg optimization of vessel speed subject to voyage time constraints. This case study illustrates how real–time predictions can be integrated with route planning to reduce underwater noise exposure while maintaining operational feasibility.

We consider a ship transiting along a prescribed route $\{\mathbf{w}_0,\mathbf{w}_1,\dots,\mathbf{w}_K\}$ where $\mathbf{w}_i$ are (latitude, longitude). Let $\mathbf{m}=(lat_m,lon_m,z_m)$ denote a marine mammal at depth $z_m$. Each leg $i$ connects waypoints $(\mathbf{w}_{i},\mathbf{w}_{i+1})$ with great–circle length
\begin{equation}
L_i \;=\; d_{\text{gc}}(\mathbf{w}_i,\mathbf{w}_{i+1}) ,
\qquad i=0,\dots,K-1,
\end{equation}
where $d_{\text{gc}}(\cdot,\cdot)$ is the haversine distance on a sphere of radius $R_\oplus$. A baseline schedule specifies a nominal speed $V_0$ (knots) and a per-leg time budget.
\begin{equation}
t_i^{(0)} \;=\; \frac{L_i}{V_0\,\kappa}, \qquad \kappa=0.514 \ \text{m\,s}^{-1}\text{ per knot}.
\end{equation}
For a source of length $L$ (m) traveling at speed $V$ (knots), the near–field source level in the one-third octave band is given by the JOMOPANS–ECHO formulation:
\begin{align}
\text{SL}(f,V,L)
&= K - 20\log_{10}\!\Big(\frac{480/V_C}{1}\Big)
    - 10 \log_{10}\!\Big(\bigl(1-\frac{f}{480/V_C}\bigr)^2 + D^2\Big) \nonumber\\
&\quad + 60\log_{10}\!\Big(\frac{V}{V_C}\Big)
      + 20\log_{10}\!\Big(\frac{L}{l_0}\Big)
      + 10\log_{10}(0.231 f),
\label{eq:jomopans_echo}
\end{align}
with constants $V_C=13.9$, $K=191$, $D=3$, $l_0=100$, and for the present case study we consider $L= 200$ m. The transmission loss for a source at $\mathbf{x}_s$ and the receptor at $\mathbf{m}$ is provided by the proposed surrogate,
\begin{equation}
\widehat{\text{TL}}(\mathbf{x}_s,\mathbf{m};f),
\end{equation}
which approximates geometric spreading, absorption, and range–dependent environmental effects. The instantaneous received level (dB re $1\,\mu$Pa) along the trajectory $\mathbf{x}_s(t)$ is
\begin{equation}
\text{RL}(t) \;=\; \text{SL}\!\big(f,V(t),L\big) \;-\; \widehat{\text{TL}}\!\big(\mathbf{x}_s(t),\mathbf{m};f\big).
\end{equation}
Sound exposure level (SEL) over an interval $\mathcal{I}$ is
\begin{equation}
\text{SEL}(\mathcal{I}) \;=\; 10 \log_{10}\!\left(\int_{\mathcal{I}} 10^{\text{RL}(t)/10}\, dt \right)\ \text{(dB)}.
\end{equation}

Each leg is optimized independently. We choose a constant leg speed $V_i\in[V_0,V_{\max}]$ to minimize the SEL per leg, subject to the schedule constraint that leg time does not exceed the baseline:
\begin{align}
\min_{V_i \in [V_0,V_{\max}]} \quad
& \text{SEL}_i \;=\; 10 \log_{10}\!\left(\int_{0}^{T_i(V_i)} 10^{\text{RL}_i(t;V_i)/10}\, dt\right), \\
\text{s.t.}\quad
& T_i(V_i) \le t_i^{(0)}, \qquad T_i(V_i)=\frac{L_i}{V_i\,\kappa}.
\end{align}
Numerically, the leg is discretized into $n_i=\lceil L_i/(V_i\,\kappa\,\Delta t)\rceil$ steps of duration $\Delta t$. Midpoint sampling along the geodesic,
\begin{equation}
\mathbf{x}_{s,i}^{(j)} \;=\; (1-\xi_j)\,\mathbf{w}_i + \xi_j\,\mathbf{w}_{i+1},
\qquad \xi_j=\frac{j+\tfrac{1}{2}}{n_i}, \quad j=0,\dots,n_i-1,
\end{equation}
gives a discrete SEL objective
\begin{equation}
\widehat{\text{SEL}}_i(V_i)
\;=\;
10\log_{10}\!\left(
\sum_{j=0}^{n_i-1} 10^{\text{RL}_i^{(j)}(V_i)/10}\, \Delta t
\right),
\qquad
\text{RL}_i^{(j)}(V_i)=\text{SL}(f,V_i,L)-\widehat{\text{TL}}(\mathbf{x}_{s,i}^{(j)},\mathbf{m};f).
\end{equation}
A bounded grid search over $V_i\in[V_0,V_{\max}]$ returns the minimizer for each leg. In this case study, we generate a grid with 200 uniformly space points. Concatenating the optimized legs produces a variable speed track $(\mathbf{x}_s(t),V(t))$ and the corresponding received-level time series on the receptor.

\begin{figure}[htbp]
    \centering
    \includegraphics[width=0.95\linewidth]{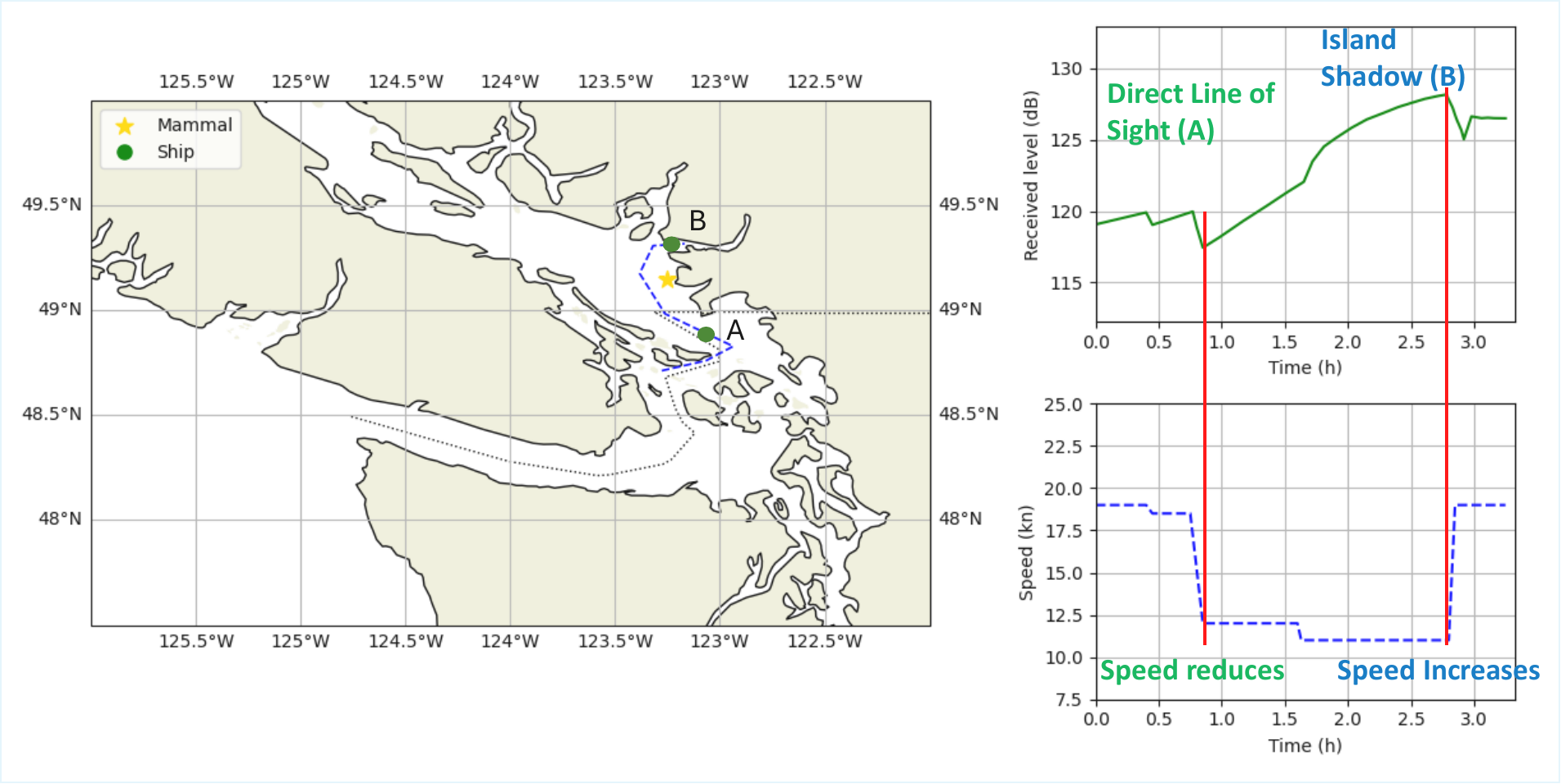}
    \caption{Operational speed optimization along a fixed route: left panel—map view with receptor (star) and ship track; top–right—received level time series; bottom–right—optimized speed profile. Per–leg speeds are chosen by minimizing discrete SEL under a per–leg time constraint.}
    \label{fig:path_opt}
\end{figure}

The optimization strategy lowers acoustic exposure at the receptor by reducing vessel speed along segments with a direct line of sight and increasing it where geometric spreading or environmental shielding attenuates the received level. Figure~\ref{fig:path_opt} shows the route of transit, the location of the receptor, and the time evolution of the received level and the speed of the vessel. The results confirm that the optimizer decreases speed when noise levels rise along the direct line of sight and increases it where propagation losses reduce exposure.
To further enhance reliability in dynamic ocean conditions, the framework is next extended to incorporate sensor assimilation, enabling the digital twin to adapt predictions in real time as new observations become available.

\subsection{Probabilistic Surrogate with Sensor Data Assimilation}
A key component of the digital twin framework is the assimilation of sparse sensor measurements, such as hydrophone recordings, into the surrogate model to improve accuracy and reduce uncertainty in prediction. Let $\mathcal{Y}_{\mathrm{pred}} = \{\widehat{\mathrm{TL}}(\mathbf{x}_j)\}_{j=1}^J$ denote the surrogate predictions at $J$ receiver locations, and let $\mathcal{Y}_{\mathrm{obs}} = \{y_k\}_{k=1}^K$ represent sensor measurements at $K \ll J$ hydrophone positions. In the variational Gaussian process setting, the residual function $f(\mathbf{z})$ admits the approximate posterior distribution given by:
\begin{equation}
    q(f) = \int p(f \mid \mathbf{u})\,q(\mathbf{u})\,d\mathbf{u}, \qquad q(\mathbf{u}) = \mathcal{N}(\mathbf{m}, \mathbf{S}),
\end{equation}
where $\mathbf{u}=f(\mathbf{Z})$ are the inducing variables at locations $\mathbf{Z}=\{\mathbf{z}_m\}_{m=1}^M$ and $\mathbf{z}_k$ are the latent embeddings of the hydrophone positions. The assimilation of sensor data corresponds to a Bayesian update of the variational posterior:
\begin{equation}
    q_{\text{post}}(f) \;\propto\; q(f)\,\prod_{k=1}^K \mathcal{N}\!\big(y_k \mid f(\mathbf{z}_k), \sigma_n^2\big),
\end{equation}
which modifies both the mean and variance of predictions in neighborhoods constrained by the observations. Explicitly, the predictive mean and variance after assimilation become:
\begin{align}
    \mu_{\text{post}}(\mathbf{z}_\star) &= \mu_{\text{GP}}(\mathbf{z}_\star) 
    + \mathbf{k}_{\star K}\left(\mathbf{K}_{KK} + \sigma_n^2 \mathbf{I}\right)^{-1}\big(\mathbf{y}_K - \mu_{\text{GP}}(\mathbf{z}_K)\big), \\
    \sigma^2_{\text{post}}(\mathbf{z}_\star) &= \sigma^2_{\text{GP}}(\mathbf{z}_\star) 
    - \mathbf{k}_{\star K}\left(\mathbf{K}_{KK} + \sigma_n^2 \mathbf{I}\right)^{-1}\mathbf{k}_{K\star},
\end{align}
where $\mathbf{k}_{\star K}$ is the cross-covariance between the test point and hydrophone embeddings, $\mathbf{K}_{KK}$ is the covariance among hydrophone embeddings, and $\mathbf{y}_K$ is the vector of sensor measurements. These expressions demonstrate that assimilation shifts the mean towards the observed values while reducing predictive variance locally.
The sensor data integration process in the digital twin cycle occurs in three stages: (i) the probabilistic surrogate provides predictions of TL fields with associated uncertainty, (ii) hydrophone measurements are integrated into the GP posterior, improving local accuracy and reducing uncertainty in the region of interest, and (iii) the updated posterior informs operational decisions such as ship speed adjustments.

\begin{figure}[htbp]
    \centering
    \includegraphics[width=1.0\linewidth]{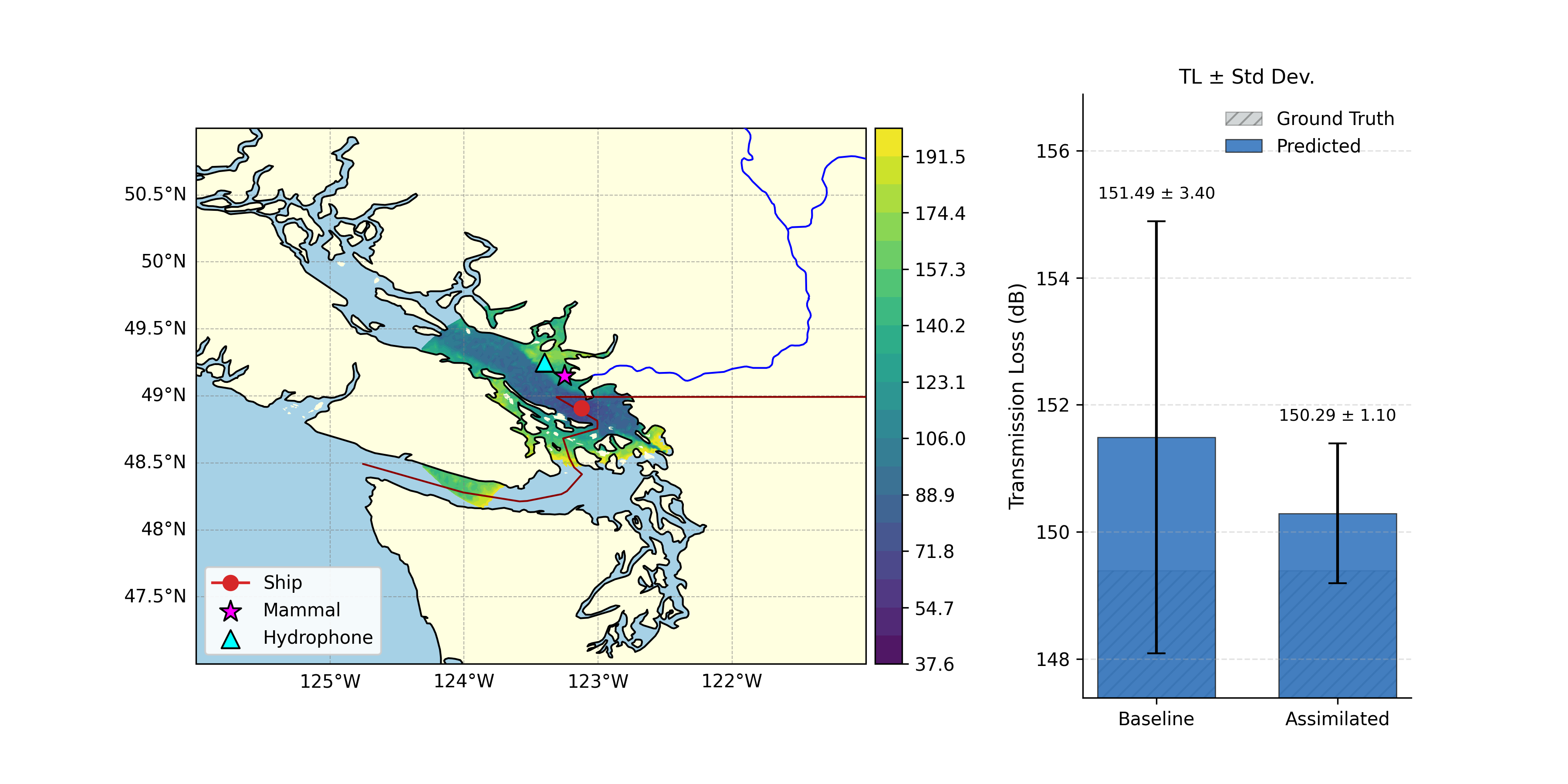}
    \caption{Illustration of surrogate update using sparse sensor measurements. 
    Left: spatial configuration showing the marine mammal (star), ship location (red circle), and hydrophone station (blue triangle). 
    Right: transmission loss prediction from surrogate (blue bar), Bellhop 3D ground truth (gray hatched bar) and predictive standard deviation before (Baseline) and after assimilation (Assimilated).}
    \label{fig:hydrophone_assimilation}
\end{figure}

To demonstrate this process, we consider an experiment in which a hydrophone is placed near a marine mammal location at $(49.25^{\circ}\,\text{N},\,123.45^{\circ}\,\text{W},\,30\,\text{m})$.
From Bellhop 3D simulations, we generate a ground truth observation $y_{\mathrm{obs}}$ at frequency $400\,\text{Hz}$ and receiver depth $30\,\text{m}$. 
The prior predictions from the probabilistic surrogate yield a mean signed error of $2.1\,\text{dB}$ in the neighborhood of the hydrophone, with predictive standard deviation $\sigma_{\text{GP}}=3.4\,\text{dB}$. 
After assimilating the single hydrophone observation data into the probablistic surrogate using the variational posterior, the mean signed error reduces to $0.9\,\text{dB}$ and the predictive standard deviation contracts to $1.1\,\text{dB}$. The effect of assimilating sparse hydrophone data is illustrated in Fig.~\ref{fig:hydrophone_assimilation}, which shows the spatial configuration of the marine mammal, ship and hydrophone station along with the reduction in mean signed error and predictive uncertainty.  
This test demonstrates that assimilation of sparse sensor data significantly improves predictive fidelity near the sensor location. In operational use, such assimilation enables the digital twin to adapt predictions dynamically, improving the reliability for decision-making in ecologically sensitive waters.

\section{Conclusion}
\label{sec:conclusion}
 In this work, we introduced a probabilistic digital twin framework for predicting transmission loss and managing operational noise in coastal waters. The framework combines a physics-informed mean function with learned feature and bathymetry encoders and a sparse variational Gaussian process residual head, providing accurate transmission loss estimates with calibrated uncertainty. By embedding physical priors into the mean function and using nonstationary embeddings derived from the encoders, the surrogate achieves improved generalization in range-dependent environments while maintaining interpretability.

Numerical experiments demonstrated that the proposed approach outperforms baseline SVGP models and hierarchical GP formulations such as DSPP and SVDKL, achieving a lower error and an improved uncertainty calibration. Large-scale case studies with Bellhop 3D simulations in the Salish Sea confirmed that the surrogate reproduces three-dimensional transmission loss fields with orders-of-magnitude acceleration relative to physics-based solvers, enabling near real-time application. Furthermore, the operational speed optimization case study illustrated how the framework can dynamically adjust vessel speed along a route to reduce noise exposure at protected receptor sites while adhering to voyage-time constraints. Assimilation experiments with sparse hydrophone data highlighted the ability of the digital twin to reduce both prediction bias and uncertainty, directly improving reliability for decision making.

Future work will focus on integrating the proposed surrogate into the MUTE-DSS framework developed in our lab \cite{venkateshwaran2025mute, venkateshwaran2024multi}. MUTE-DSS is a digital twin-based decision support system that combines near-field semi-empirical models with 3D ray tracing for far-field propagation, coupled with data-driven models of Southern resident killer whale distribution, and performs two-stage voyage optimization using sampling-based planning and genetic algorithms. Incorporating the proposed surrogate in place of the computationally expensive ray-tracing component would allow MUTE-DSS to scale to larger fleets and longer planning horizons, while retaining uncertainty-aware predictions of transmission loss. This integration has the potential to enable adaptive, real-time decision support for ship routing and speed profiling that minimizes underwater radiated noise exposure to marine mammals at an operational scale and can be further extended to multi-sensor networks and fleet-level optimization for sustainable maritime operations.

\section*{Acknowledgment}
The present study is supported by Mitacs, Transport Canada, and Clear Seas through the Quiet-Vessel Initiative (QVI) program. 
The authors would like to express their gratitude to Dr. Paul Blomerous and Ms. Tessa Coulthard for their valuable feedback and suggestions.  We also acknowledge that the GPU facilities at the Digital Research Alliance of Canada clusters were used for the training of our deep learning models.

\bibliography{references}


\end{document}